\documentclass[10pt,twocolumn,letterpaper]{article}

\usepackage{cvpr}              

\usepackage[dvipsnames]{xcolor}

\usepackage{color, xcolor}
\usepackage{soul}
\usepackage{ulem}
\usepackage{multirow}
\usepackage{multicol}
\usepackage{makecell}
\usepackage{array}
\usepackage{dsfont}
\definecolor{cvprblue}{rgb}{0.21,0.49,0.74}
\usepackage[pagebackref,breaklinks,colorlinks,citecolor=cvprblue]{hyperref}

\def\paperID{22} 
\def\confName{3DV\xspace}
\def\confYear{2025\xspace}

\newcommand{\myFigRef}[1]{Fig.~\ref{#1}}
\newcommand{\mySecRef}[1]{Sec.~\ref{#1}}

\definecolor{myPurple}{rgb}{0.4, .0, .8}
\definecolor{myGreen}{rgb}{0, .8, .3}
\definecolor{myRed}{rgb}{0.8, .2, .2}
\definecolor{myOrange}{rgb}{0.7, 0.45, 0.2}
\definecolor{myBlue}{rgb}{.0, .0, 1.0}
\definecolor{myBlue2}{rgb}{.0, .0, 0.5}
\definecolor{myBlack}{rgb}{.0, .0, 0.0}

\title{Ctrl-Room: Controllable Text-to-3D Room Meshes Generation with Layout Constraints}

\author{
Chuan Fang$^{1,2}$\footnotemark[1]~, Yuan Dong$^{3}$\footnotemark[1]~, Kunming Luo$^{1,2}$, Xiaotao Hu$^{1,2}$,  Rakesh Shrestha$^{4}$, Ping Tan$^{1,2}$\footnotemark[2]~ \\
 $^{1}$ Hong Kong University of Science and Technology $^{2}$ LightIllusion, China.\\ $^{3}$ Alibaba Group $^{4}$Simon Fraser University, Canada.\\
{\tt\small $^{1}$cfangac@connect.ust.hk, \tt\small $^{2}$dy283090@alibaba-inc.com,  $^{3}$pingtan@ust.hk} \\
{\tt\small \url{https://fangchuan.github.io/ctrl-room.github.io/}}
}

\begin{document}
 \maketitle
 \begin{abstract}
Text-driven 3D indoor scene generation is useful for gaming, film industry, and AR/VR applications. However, existing methods cannot faithfully capture the scene layout based on text descriptions, nor do they allow flexible editing of individual objects in the room.
To address these problems, we present Ctrl-Room, which can generate convincing 3D rooms with designer-style layouts and high-fidelity textures from just a text prompt. Our key insight is to separate the modeling of layouts and appearance.
Our proposed method consists of two stages: a Layout Generation Stage and an Appearance Generation Stage. The Layout Generation Stage trains a text-conditional diffusion model to learn the layout distribution with our holistic scene code parameterization. Next, the Appearance Generation Stage employs a fine-tuned ControlNet to produce a vivid panoramic image of the room guided by the 3D scene layout, then further upgrades to a panoramic NeRF model. 
Benefiting from the scene code parameterization, we can easily edit the generated room model through our mask-guided editing module, without expensive edit-specific training. Extensive experiments on the Structured3D dataset demonstrate that our method outperforms existing methods in producing more reasonable, view-consistent, and editable 3D rooms from text prompts. 
\end{abstract}    
 \section{Introduction}
\label{sec:intro}
High-quality 3D indoor scenes play a crucial role across a wide array of applications, ranging from interior design and video games to simulators for embodied AI.
Traditionally, indoor scenes are crafted manually by professional artists, which is both time-consuming and costly. 
Recent advancements in generative models~\cite{poole2022dreamfusion,chen2023fantasia3d,lin2023magic3d,seo2023let} have attempted to simplify the creation of 3D models from textual descriptions, 
However, extending this capability to text-driven 3D indoor scene generation presents unique challenges as they exhibit strong semantic layout constraints, such as neighboring walls are perpendicular and the TV set often faces a sofa, that are more complicated than objects.

\begin{figure}
 \vspace{-0.3in}
	\centering
	\subfloat[Comparison with Text2Room~\cite{hollein2023text2room} and MVDiffusion~\cite{tang2023mvdiffusion}.]{
		\includegraphics[width=\linewidth]{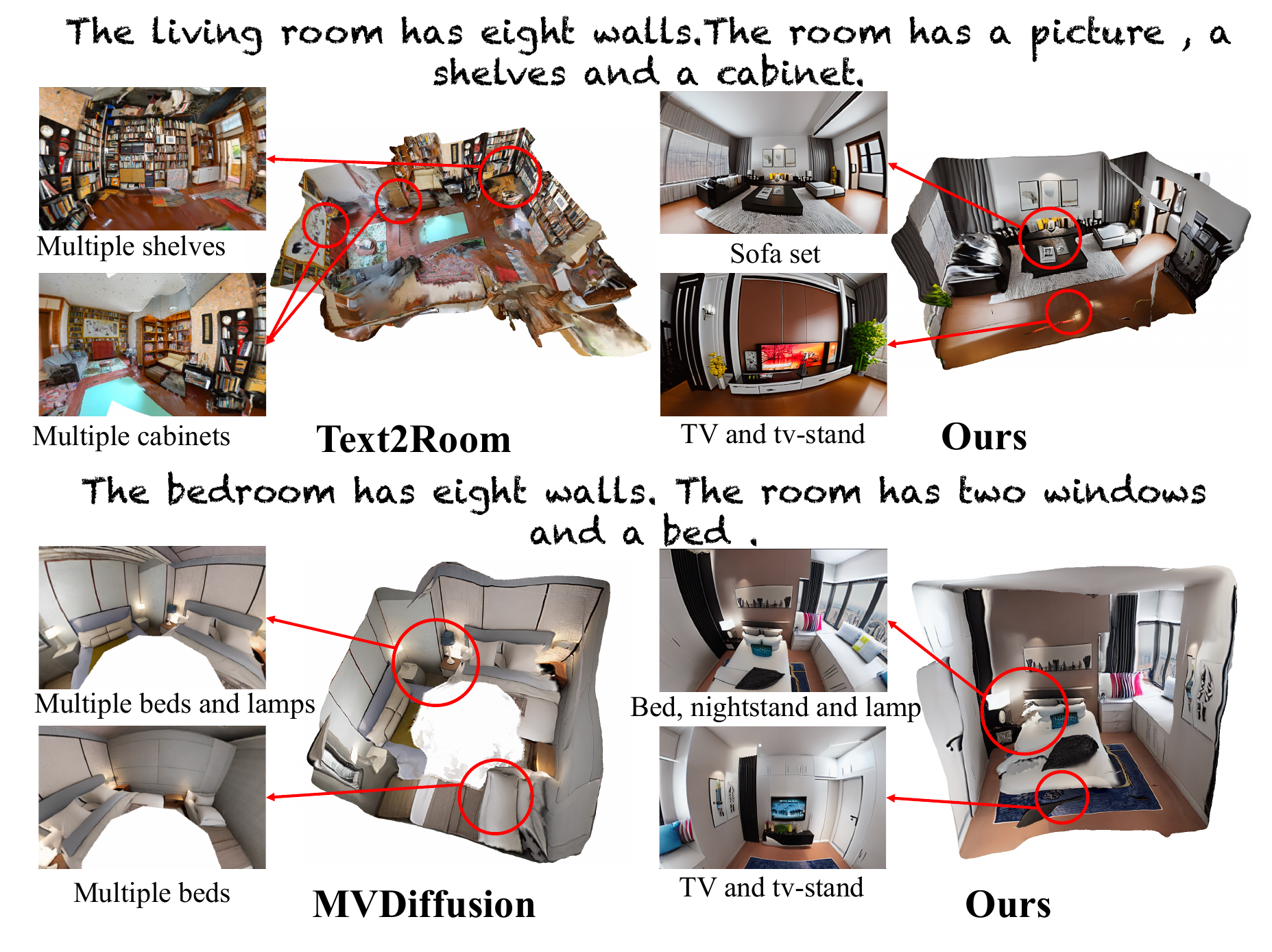}
            \vspace{-0.1in}
		\label{fig:teasers_a}
        }\\
	\centering
	\subfloat[Flexible editing by instruction or mouse clicks.]{
		\includegraphics[width=\linewidth]{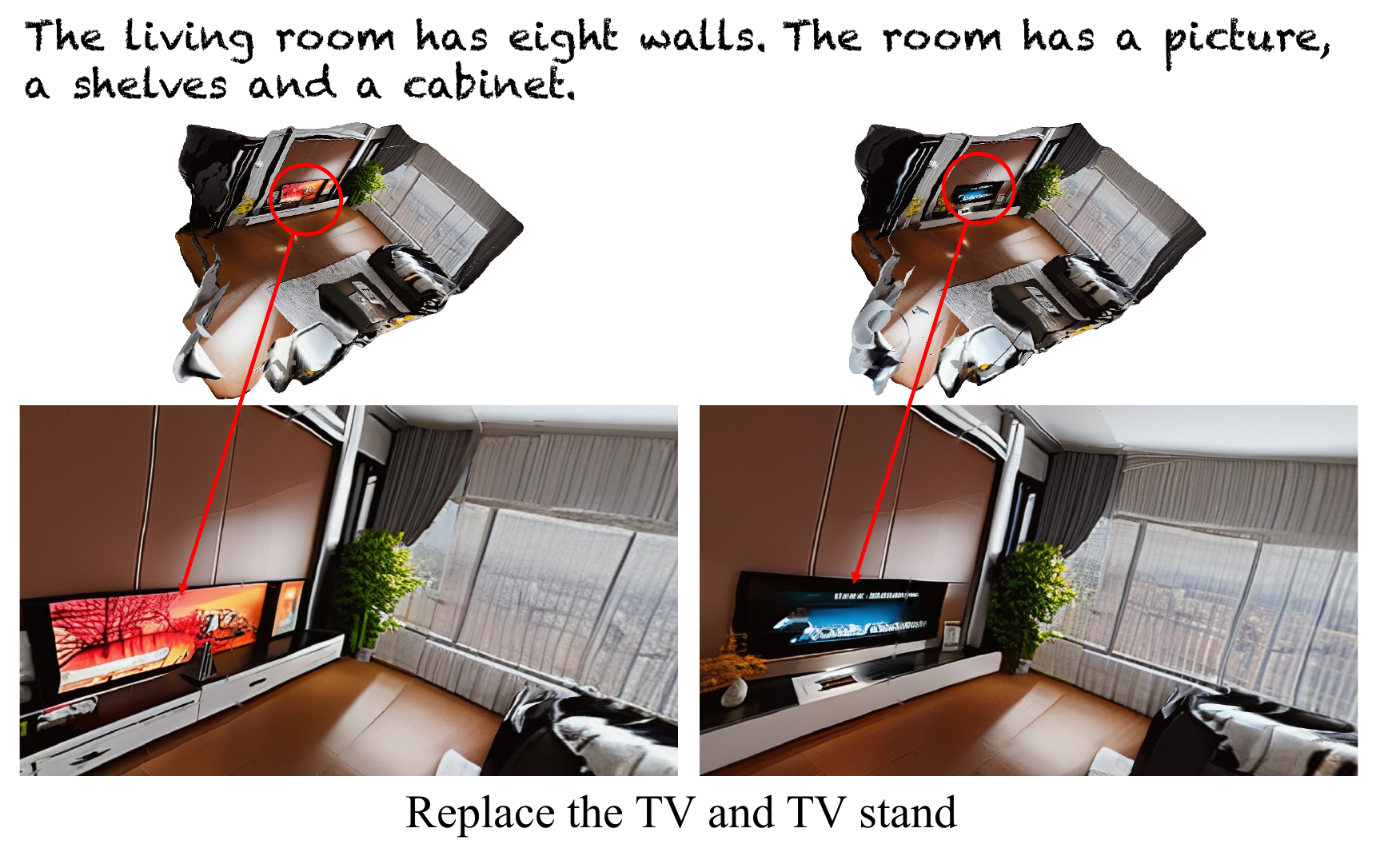}
            \vspace{-0.1in}
		\label{fig:teasers_b}}
        \vspace{-0.1in}
	\caption{We present \textit{Ctrl-Room} to achieve fine-grained textured 3D indoor room generation and editing. (a) compared with the Text2Room~\cite{hollein2023text2room} and MVDiffusion\cite{tang2023mvdiffusion}, Ctrl-Room can generate rooms with more plausible 3D structures. (b) Ctrl-Room supports flexible editing. Users can replace furniture items or change their positions easily. 
	}
 \label{fig:teaser_a}
 \vspace{-0.3in}
\end{figure}

\noindent Existing text-driven 3D indoor scene generation approaches, such as Text2-Room~\cite{hollein2023text2room} and Text2NeRF~\cite{zhang2023text2nerf}, are designed with an incremental framework. They create 3D indoor scenes by incrementally generating different viewpoints frame-by-frame and reconstructing the 3D mesh of the room from these sub-view images. However, these approaches often fail to model the global layout of the room, resulting in unconvincing results. As shown in the first row of ~\cref{fig:teaser_a} (a), the result of Tex2Room exhibits repetitive objects, e.g. several cabinets in a living room, and does not follow the furniture layout patterns. We refer to this problem as the \textit{`Penrose Triangle problem'}, where a generated scene has plausible 3D structures everywhere locally but lacks global consistency. Furthermore, prior approaches do not offer user-friendly interaction, as the resulting 3D geometry and textures are not editable.  Other method~\cite{lin2019coco, lin2021infinitygan, shum2023conditional,tang2023mvdiffusion} represent the scene as a panorama image and generate it from a text prompt. However, these works cannot guarantee reasonable scene layouts. As shown on the middle row of \cref{fig:teaser_a} (a), a bedroom generated by MVDiffusion~\cite{tang2023mvdiffusion} contains multiple beds, which violates room layout priors.


\noindent To address these shortcomings, we propose a novel two-stage method to generate a high-fidelity and editable 3D room. The key insight is to separate the generation of 3D geometric layouts from that of visual appearance, which allows us to better capture the room layout and achieve vivid textures at the same time. 
In the first stage, from text input, our method creates plausible scene layouts with various furniture types and positions. Unlike previous scene synthesis methods~\cite{tang2023diffuscene, paschalidou2021atiss} that only focus on the furniture arrangement, our approach further considers walls with doors and windows, which play an essential role in the layout. To achieve this goal, we parameterize the room by a holistic scene code, which represents a room as a set of objects. Each object is represented by a vector capturing its position, size, semantic class, and orientation. Based on our compact parameterization, we design a diffusion model to learn the 3D room layout distribution from the Structured3D dataset~\cite{zheng2020structured3d}.

\noindent Our method then generates the room appearance with the guidance of the 3D room layout. We first generate a panorama using a text-to-image latent diffusion model, then iteratively upgrade the generated images to a NeRF model and generate additional novelty view panorama images. During the panorama generation, unlike previous text-to-panorama works~\cite{tang2023mvdiffusion, chen2022text2light}, our method explicitly enforces scene layout constraints and guarantees plausible 3D room structures and furniture arrangement. To achieve this goal, we convert the 3D layout synthesized in the first stage into a semantic segmentation map and feed it to a fine-tuned ControlNet~\cite{zhang2023adding} model to create the panorama image. We also use this layout information to estimate scene depth and inpaint missing regions at novel viewpoints.

\noindent Benefiting from the separation of layout and appearance, our method enables flexible editing on the generated 3D room. The user can replace or modify the size and position of furniture items, e.g. replacing the TV and TV stand as in \cref{fig:teaser_a} (b). Our method can update the room according to the edited room layout through our mask-guided editing module without expensive edit-specific training. The updated room appearance maintains consistency with the original version while satisfying the user's edits. 

\noindent The main contributions of this paper are summarized as:

\begin{itemize}
    \item 
    To address the Penrose Triangle Problem, we design a two-stage method for 3D room generation from pure text input, which separates the geometric layout generation and appearance generation. In this way, our method can better capture the scene layout constraints in real-world data and produce a vivid appearance simultaneously. 
    \item
    Within the separated layout and appearance generation, we introduce novel techniques, including holistic scene code parametrization, layout-guided panorama generation, layout-guided panoramic NeRF, and a mask-guided editing module to achieve high-quality and flexible 3D room generation.
   \item
       Qualitative and quantitative experiments confirm that our method excels in producing more realistic and editable 3D rooms compared to existing approaches.
    
\end{itemize}

\section{Related Work}
\label{sec:related_work}

\begin{figure*}[t]
    \vspace{-0.3in}
    \centering
    \includegraphics[width=\linewidth]{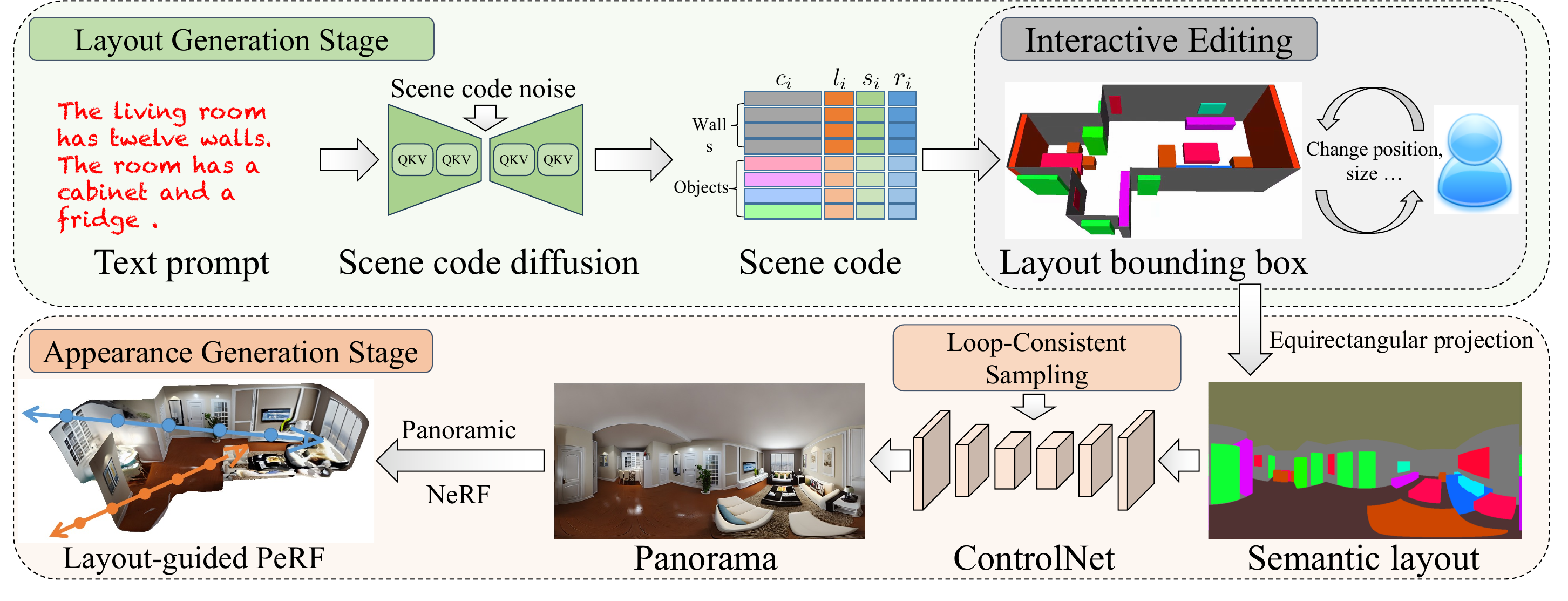}
    \vspace{-0.2in}
    \caption{Framework overview. In Layout Generation Stage, we synthesize a scene code from the text input and convert it to a 3D bounding box representation to facilitate editing. In Appearance Generation Stage, we project the bounding boxes into a semantic segmentation map to guide the panorama synthesis. The panorama is then reconstructed into a panoramic NeRF (PeRF)\cite{wang2023perf}model with layout guidance.}
    \label{fig:system_framework}
    \vspace{-0.2in}
\end{figure*}

\subsection{Text-based 3D Object Generation}
Early methods employ 3D datasets to train generative models. Text2Shape~\cite{chen2019text2shape} learns a feature representation from paired text and 3D data and uses GAN to generate 3D shapes from the text.  Point-E~\cite{nichol2022point} and Shap-E~\cite{Jun2023ShapEGC} enlarge the scope of the training dataset and employ a latent diffusion model~\cite{rombach2022high} for object generation. However, 3D datasets are scarce, which makes these methods difficult to scale. More recent methods~\cite{nichol2022point, poole2022dreamfusion, lin2023magic3d, wang2023score, chen2023fantasia3d, wang2023prolificdreamer} exploit the powerful 2D text-to-image diffusion models~\cite{rombach2022high, 
 saharia2022photorealistic} for 3D model generation. Typically, these methods generate one or multiple 2D images in an incremental fashion and optimize the 3D model accordingly. DreamFusion~\cite{poole2022dreamfusion} introduces a loss based on probability density distillation and optimizes a randomly initialized 3D model through gradient descent. Magic3D~\cite{lin2023magic3d} uses a coarse model to represent 3D content and accelerates it using a sparse 3D hash grid structure. 
 To alleviate over-saturation and low-diversity problems, ProlificDreamer~\cite{wang2023prolificdreamer} models and optimizes the 3D parameters through variational score distillation. However, these methods are limited to 3D object generation and cannot be directly extended to 3D scene generation which has additional layout constraints.

\subsection{Text-based 3D Room Generation}
\paragraph{Room Layout Synthesis} Layout generation has been greatly boosted by transformer-based methods. LayoutTransformer~\cite{gupta2021layouttransformer} employs self-attention to capture relationships between elements to accomplish layout completion. ATISS~\cite{paschalidou2021atiss} proposes an autoregressive transformer to generate proper indoor scenes with only the room type and floor plan as the input. 
DiffuScene~\cite{tang2023diffuscene} and InstructScene~\cite{lin2024instructscene} model a union of furniture as a fully connected scene graph and propose a diffusion model to sample physically plausible scenes. While these methods generate reasonable furniture layouts, they do not consider the walls, doors, and windows which are crucial in the furniture arrangement. Thus they do not always generate realistic indoor environments.

\paragraph{Panoramic Image Generation}
Another line of work~\cite{lin2019coco, lin2021infinitygan, shum2023conditional} represent an indoor scene by a panorama image without modeling 3D shapes. These methods enjoy the benefits of abundant training data and produce vivid results. COCO-GAN~\cite{lin2019coco} produces a set of patches and assemble them into a panoramic image. InfinityGAN~\cite{lin2021infinitygan} uses the information of two patches to generate the parts between them, to finally obtain a panoramic image. ~\cite{shum2023conditional} proposes a 360-aware layout generator to produce furniture arrangements and uses this layout to synthesize a panoramic image based on the input scene background. MVDiffusion~\cite{tang2023mvdiffusion} simultaneously generates multi-view perspective images and proposes a correspondence-aware attention block to maintain multi-view consistency, and then transfers these images to a panorama. These methods might suffer from incorrect room layout since they do not enforce layout constraints. Furthermore, the results of these methods cannot be easily edited, e.g. resizing or moving furniture around, because they do not maintain an object-level representation. 

\paragraph{3D Room Generation}
GAUDI~\cite{bautista2022gaudi} generates immersive 3D indoor scenes rendered from a moving camera. It disentangles the 3D representation and camera poses to ensure the consistency of the scene during camera movement.
CC3D~\cite{bahmani2023cc3d} proposes a 3D-aware GAN for multi-object scenes conditioned on a single semantic layout image and is trained using posed multi-view RGB images. Another related line of work~\cite{song2023roomdreamer,yang2023dreamspace,schult2023controlroom3d} deals with retexturizing a given 3D scene. They employ 2D diffusion models to stylize and further improve the given geometry.
Text2Room~\cite{hollein2023text2room} incrementally synthesizes nearby images with a 2D diffusion model and recovers its depth maps to assemble into a 3D room mesh. Unfortunately, it cannot handle the geometric and textural consistency among the images, resulting in the \textit{`Penrose Triangle problem'}. In our method, we take both geometry and appearance into consideration and create a more geometrically plausible 3D room. 
A concurrent work~\cite{schult2023controlroom3d} also guides the 3D room mesh generation by leveraging the user-input scene layouts. In contrast, our method is capable of synthesizing professional designer-style layouts solely from text prompts.

 \section{Method}
\label{sec:method}

\begin{figure*}[t]
    \vspace{-0.2in}
    \centering
    \includegraphics[width=0.9\linewidth]{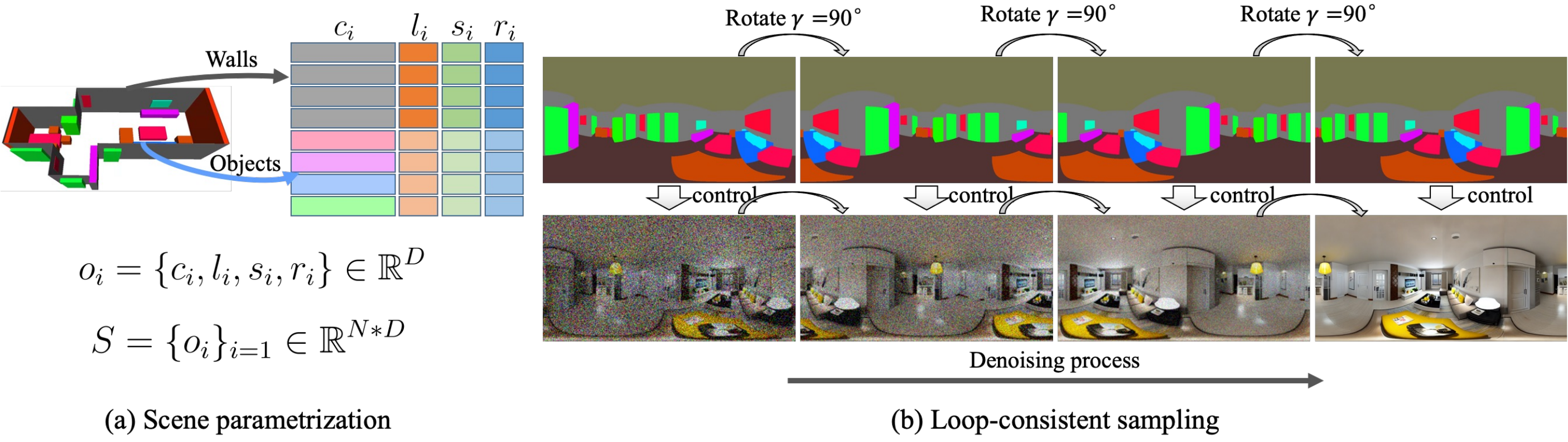}
    \vspace{-0.15in}
    \caption{(a) A 3D scene $S$ is represented by its scene code $x_0 = \{o_i\}_{i=1}^{N}$, where each wall or furniture item $o_i$ is a row vector storing attributes like class label $c_i$, location $l_i$, size $s_i$, orientation $r_i$. (b) During the denoising process, we rotate both the input semantic layout panorama and the denoised image for $\gamma$ degree at each step. Here we take $\gamma=90^\circ$ for example.}
    \label{fig:method_detail}
    \vspace{-0.2in}
\end{figure*}

\noindent In order to achieve text-based 3D indoor scene generation, we propose \textbf{Ctrl-Room}. We first generate the room layout from an input text and then generate the room appearance represented by panoramic images according to the layout, followed by layout-guided panoramic NeRF~\cite{wang2023perf} to generate the final 3D room. This mechanism solves the \textit{Penrose Triangle Problem} to generate physically plausible 3D rooms, while also enabling users to edit the scene layout interactively. The overall framework of our method is depicted in ~\myFigRef{fig:system_framework}, which consists of two stages: the Layout Generation Stage and the Appearance Generation Stage. 
In the Layout Generation Stage, we parameterize the indoor scene with a holistic scene code and design a diffusion model to learn its distribution.
Once the holistic scene code is generated from text, we recover the room as a set of orientated bounding boxes of walls and objects. Note that users can edit these bounding boxes by adjusting their semantic types, positions, or scales, enabling the customization of 3D room generations. 
In the Appearance Generation Stage, we obtain an RGB panorama through a conditioned image diffusion model to represent the room texture. Specifically, we project the generated layout bounding boxes into a semantic segmentation layout. We then fine-tune a pre-trained ControlNet~\cite{zhang2023adding} model to generate an RGB panorama from the input semantic layout. To ensure loop consistency, we propose a loop-consistent sampling during the inference process.
Finally, we integrate the layout and the panorama, then generate a full 3D room through the layout-guided panoramic NeRF module~\cite{wang2023perf}. This module progressively inpaints panoramas at new viewpoints using the fine-tuned ControlNet. To extract meshes from reconstructed NeRF, we render depth maps of the new views and utilize truncated signed distance fusion (TSDF) to obtain the final mesh.

\subsection{Layout Generation Stage}
\label{sec:method_layout_gen}




\noindent{\bf Scene Code Definition.}
Different from previous methods ~\cite{paschalidou2021atiss,tang2023diffuscene}, we consider not only furniture but also walls, doors, and windows to define the room layout. We employ a unified encoding of various objects. Specifically, given a 3D scene $\mathcal{S}$ with $m$ walls and $n$ furniture items, we represent the scene layout as a holistic scene code $\mathbf{x_0}=\{ \mathbf{o_i} \}^{N}_{i=1}$, where $N=m+n$. We encode each object $o_j$ as a node with attributes including center location $l_i \in \mathbb R^3$, size $s_i \in \mathbb R^3$, orientation $r_i \in \mathbb R$, class label $c_i \in \mathbb R^{C}$. The concatenation of these attributes characterizes each node as $\mathbf o_i = [c_i, l_i, s_i, r_i]$. 
As can be seen in Fig.~\ref{fig:method_detail} (a), we represent a scene layout as a tensor $\mathbf{x_0} \in \mathbb{R}^{N\times D}$, where $D$ is the attribute dimension of a node. In all the data, we choose the normal direction of the largest wall as the `main direction'. For other objects, we take the angles between their front directions and the main direction as their rotations.  We use the one-hot encoding to represent their semantic types.

\noindent{\bf Scene Code Diffusion.}  With the scene code definition, we build a diffusion model to learn its distribution.
A scene layout is a point in $\mathbb{R}^{N \times D}$. The forward diffusion process is a discrete-time Markov chain in $\mathbb{R}^{N \times D}$. Given a clean scene code $\mathbf x_0$, the diffusion process gradually adds Gaussian noise to $\mathbf x_0$, until the resulting distribution is Gaussian, according to a pre-defined, linearly increased noise schedule $\beta_1, ..., \beta_T$:
\begin{equation}
    q(\mathbf{x_t}|\mathbf{x_0}) :=  \mathcal{N}(\mathbf{x_t}; \sqrt{\bar \alpha_t} \mathbf{x_0}, (1-\sqrt{\bar \alpha_t}) \mathbf I) 
\label{eq:scene_graph_diffusion1}
\end{equation}
where $\alpha_t := 1 - \beta_t$ and $\bar{\alpha_t} := \prod_{r=1}^{t} \alpha_r$ define the noise level and decrease over the timestep $t$.

\noindent The denoising network is trained to reverse the above process by minimizing the training objectives which includes the denoising objective $\mathcal{L_{\rm denoise}}$ and a regularization term $\mathcal L_{\rm physical}$ to penalize the penetration among objects and walls as follows,
\setlength{\abovedisplayskip}{-1pt}
\setlength{\belowdisplayskip}{-1pt}    
\begin{eqnarray}
    \mathcal{L} & = & \mathcal{L_{\rm denoise}} + \mathcal L_{\rm physical}, \\
    \mathcal{L_{\rm denoise}} & = & \mathbf{E}_{\mathbf{x_0}, t, y, \mathbf{\epsilon}} {\left \| \mathbf{\epsilon} - \epsilon_{\theta}(x_t,t, y)  \right \|^2}, \\
    \mathcal L_{\rm physical} & = & \sum_{t=1}^{T} w_t \ast (\mathcal L_{\rm w-o} + \mathcal L_{\rm o-o}). 
\label{eq:layout_diffusion_physical_objective}
\end{eqnarray}
\noindent where $\epsilon_{\theta}$ is the noise estimator which aims to find the noise $\mathbf \epsilon $ added into the input $x_0$. Here, $y$ is the text embedding of the input text prompts. The hyperparamter $w_t$ is set to $\mathbf{\bar {\alpha}_t}* 0.1$.
$\mathcal L_{\rm w-o}$ is the physical violation loss between walls and objects. We adopt the 3D IoU loss $\mathcal L_{\rm o-o}$ in DiffuScene to avoid intersection between furniture.

\noindent The denoising network $\mathbf{\epsilon_{\theta}}$ takes the scene code $\mathbf{x_t}$, text prompt $y$, and timestep $t$ as input, and denoises them iteratively to get a clean scene code $\mathbf{\hat x}_0$. Please refer to appendix Sec.1 for the details of our $\mathcal L_{\rm w-o}$ and denoising network. 

\subsection{Appearance Generation Stage}
\label{sec:method_appearance_gen}
Given an indoor scene layout, we seek to generate the 3D textured room model. We achieve this goal by generating panoramic images and reconstructing a panoramic NeRF (PeRF) model from these panoramas. 
During the panorama generation, instead of incrementally generating multi-view images like~\cite{hollein2023text2room}, we generate the entire panorama at once. We utilize ControlNet~\cite{zhang2023adding} to generate a high-fidelity panorama conditioned by the 3D scene layout. 

\subsubsection{Layout-guided Panorama Generation}

\noindent{\bf Fine-tuning ControlNet.} ControlNet controls the image generation of Stable Diffusion~\cite{rombach2022high} model by an extra 2D input. To condition ControlNet on the scene layout, we convert the bounding box representation into a 2D semantic layout panorama through equirectangular projection. In this way, we get a pair of RGB and semantic layout panoramic images for each scene. However, the pre-trained ControlNet-Segmentation~\cite{ControlNetGithubModel} is designed for perspective images, and cannot be directly applied to panoramas. Thus, we fine-tune it with our pairwise RGB-Semantic layout panoramas on the Structured3D~\cite{zheng2020structured3d}.
As the volume of Structured3D is limited, we apply several augmentation techniques for the training data, including standard left-right flipping, horizontal rotation, and Pano-Stretch~\cite{sun2019horizonnet}. 

\noindent{\bf Loop-consistent Sampling.} A panorama should be loop-consistent. In other words, its left and right should be seamlessly connected. Although the horizontal rotation in data augmentation may improve the model’s implicit understanding of the expected loop consistency, it lacks explicit constraints and might still produce inconsistent results. 
Therefore, we propose an explicit loop-consistent sampling mechanism in the denoising process of the latent diffusion model. 
As shown in Fig.~\ref{fig:method_detail} (b), we rotate both the input layout panorama and the denoised image by $\gamma$ degree in the sampling process, which applies explicit constraints for the loop consistency during denoising. A concurrent work~\cite{wu2023ipo} also uses a similar method for panoramic outpainting. More qualitative results in supplementary Fig.8 and Fig.9 verify that our simple loop-consistent sampling method achieves good results without introducing additional learnable parameters. 

\begin{figure}
\vspace{-0.3in}
\centering
\includegraphics[width=\linewidth]{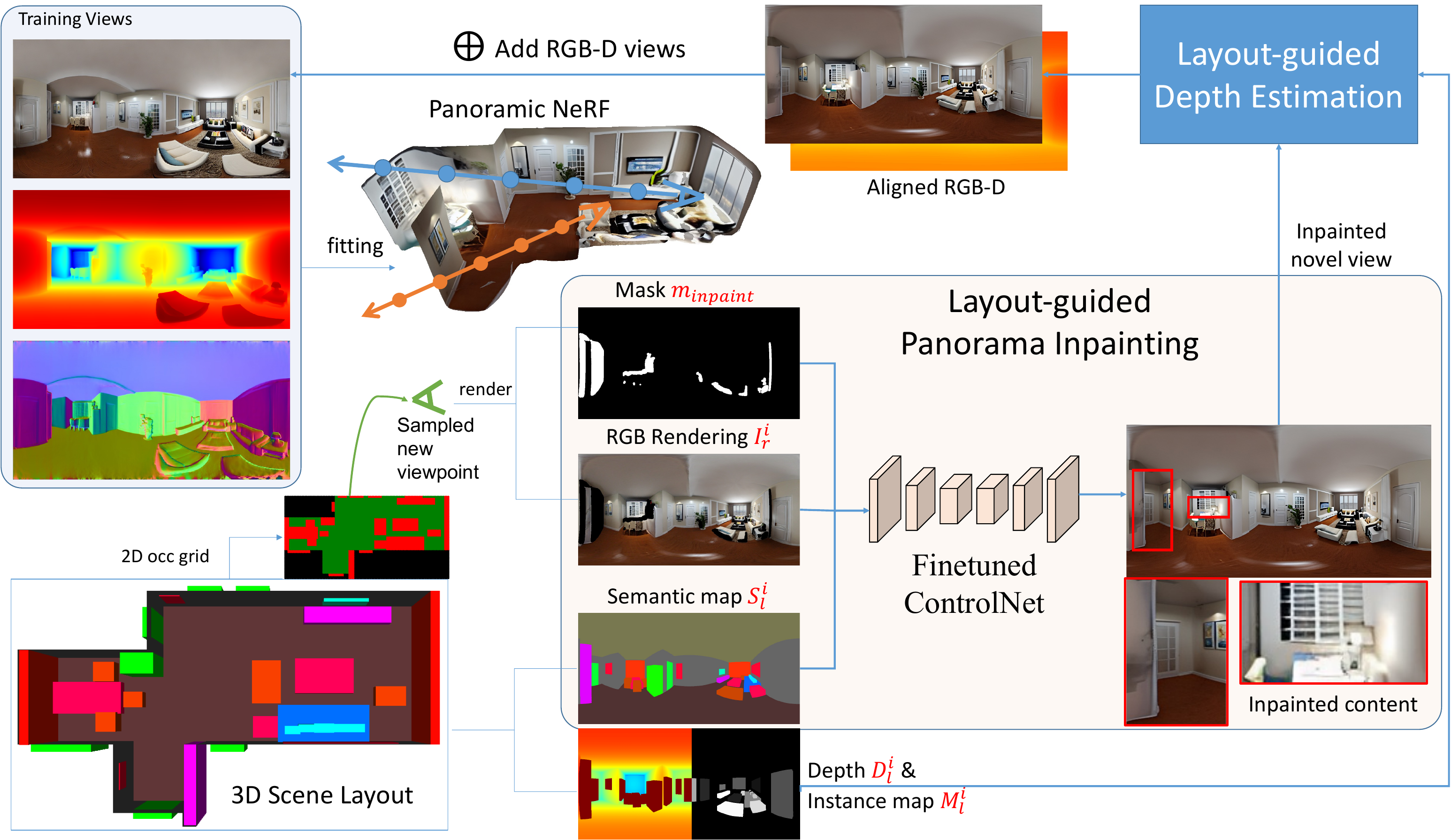}
\vspace{-0.25in}
\caption{The Layout-guided PeRF takes the input panorama, aligned depth map, and normal map as initialization. Then, a progressive inpainting module is introduced to generate consistent panoramic images at sampled novel views. The progressive inpainting module consists of the layout-guided panorama inpainting and the layout-guided depth estimation module. The final RGB-D panoramic pairs are included as training views to finetune PeRF~\cite{wang2023perf}.
}
\label{fig:method_layout_guided_perf}
\vspace{-0.2in}
\end{figure}

\subsubsection{Layout-guided PeRF Generation}
\noindent Since a single panorama is only a partial observation of a scene up to occlusions, lifting a single view into a 3D room becomes complex.  Fortunately, our generated layout provides valuable geometric and semantic information to lift the 2D panorama into a 3D model. 
We propose the layout-guided PeRF, which upgrades the generated panorama aforementioned to a 3D panoramic NeRF~\cite{wang2023perf}, enabling multi-view consistent panorama generations guided by the scene layout.
Specifically, we start with the layout-guided depth estimation, which recovers the depth map using method~\cite{yun2023egformer} and then aligns it to the 3D scene layout leveraging its geometric information. This step corrects the biased depth prediction in the background (wall, ceiling, floor) and preserves objects' shape in the foreground. 

\noindent Then, we fit our layout-guided PeRF as illustrated in \myFigRef{fig:method_layout_guided_perf}. 
Specifically, we initialize the scene NeRF with the panorama $I^0$, the aligned depth map $D^*$, and the normal map $N^*$. We sample new viewpoints in the occupancy grid that do not conflict with the initial furniture arrangement. At the i-th novel view, we render semantic map $S^i_l$, depth map $D^i_l$, and instance map $M^i_l$ from the scene layout, these are then combined with the panoramic rendering $I^i_r$ and inpainting mask $\mathbf m_{\rm inpaint}$ obtained from the NeRF and fed to the layout-guided panorama inpainting module to generate the novel view panorama. Using our fine-tuned ControlNet, it achieves training-free panoramic inpainting, which replaces pixels outside the inpainting mask $\mathbf m_{\rm inpaint}$ with $I^i_r$ and fill $\mathbf m_{\rm inpaint}$ based on the semantic map $S^i_l$.
Subsequently, after generating the novel view image, we apply the layout-guided depth estimation and include it as training views for PeRF following their framework~\cite{wang2023perf}.
More details and results can be found in the appendix Sec.2.

\begin{figure*}
    \vspace{-0.3in}
    \centering
    \includegraphics[width=1.0\linewidth] {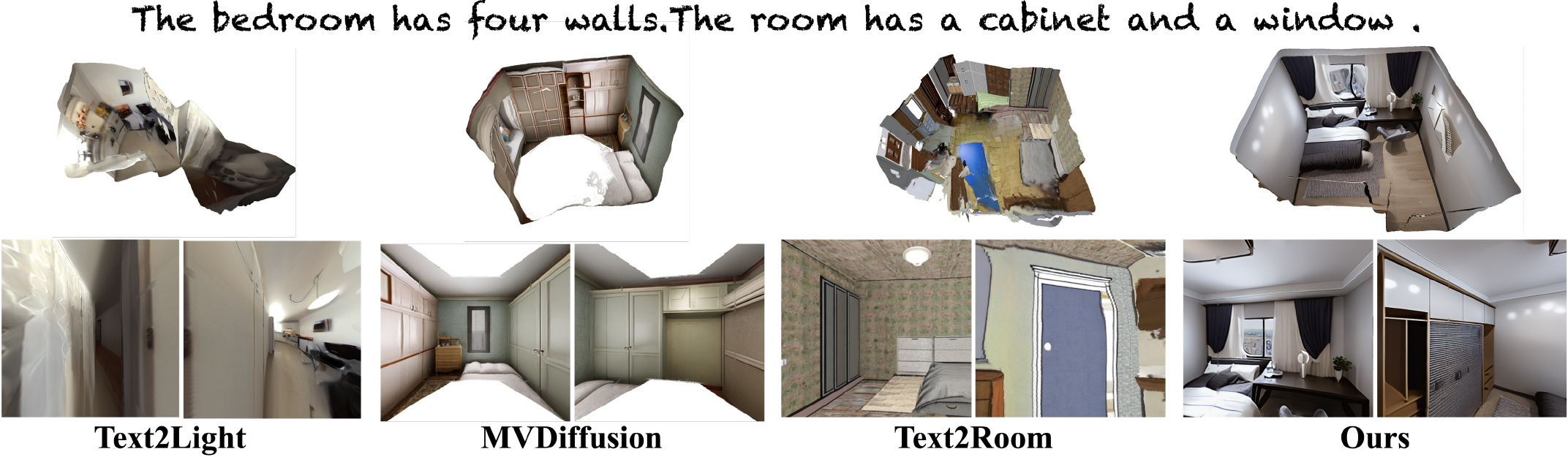}
    \vspace{-0.25in}
    \caption{Qualitative comparison with previous works. For each method, we show a textured 3D mesh in the first row and two rendered images in the second row.
    }
    \label{fig:exp_qual_comp}
    \vspace{-0.2in}
\end{figure*}

\subsection{Mask-guided Editing}
\label{sec:method_mask_guied_editing}

A user can modify the generated 3D room by changing the position, semantic class, and size of object bounding boxes. 
The editing should achieve two goals, i.e. altering the content according to the user's input, and maintaining appearance consistency of the scene objects. 
We propose a mask-guided image editing, including inpainting step and optimization step as illustrated Fig.6 in supplementary file. The inpainting step fills in the modified area while preserving the rest of the panoramic image. The optimization step focuses on keeping the furniture's appearance unchanged before and after movement and scaling operations. 

\noindent We explain our method by taking the example in Fig.6 in supplementary file, where a chair's position is moved. We denote the semantic panorama from the edited scene as $S_{\rm edited}$, then we derive the guidance masks based on its difference from the original one $S_{\rm ori}$. The source mask $\mathbf m_{\rm src}$ shows the position of the original chair, and the target mask $\mathbf m_{\rm tar}$ indicates the location of the moved chair, and the inpainting mask $\mathbf m_{\rm inpaint} = \{m | m \in \mathbf m_{\rm src} \; {\rm and} \; m \notin \mathbf m_{\rm tar} \}$ is the unoccluded region.
We use $\mathbf{x^{{\rm ori}}_{0}}$ to denote the original image. During the inpainting step, we replace pixels outside the inpainting mask $\mathbf m_{\rm inpaint}$ with $\mathbf{x^{{\rm ori}}_{t}}$ and store $\mathbf m_{\rm inpaint}$ based on the edited semantic panorama $S_{\rm edited}$. This straightforward approach ensures that the region outside the mask remains unchanged and the area inside the mask is accurately inpainted.
In the optimization step, drawing inspiration from DIFT~\cite{tang2023emergent}, which has shown that learned features from the diffusion network enable strong semantic correspondence, we ensure consistency between the original and moved furniture by requiring their latent features to be consistent. For more details of the Inpainting and Optimization Step, please refer to our supplementary file Sec.3.

 \section{Experiments}
\label{sec:exp}
We evaluate Ctrl-Room on three tasks: layout generation, panorama generation, and 3D Room generation. For those panorama generation methods~\cite{chen2022text2light, tang2023mvdiffusion}, we recover its depth map using method~\cite{yun2023egformer} to reconstruct a textured mesh through Possion reconstruction~\cite{kazhdan2006poisson} and MVS-texture~\cite{waechter2014let}. We first describe the experimental settings and then validate our method by comparing it with previous methods quantitatively and qualitatively. We further show various scene editing results to demonstrate the flexible control of our method.

\subsection{Experiment Setup}
\label{sec:exp_setup}
\textbf{Dataset:} We train and evaluate our method on the 3D indoor scene dataset Structured3D~\cite{zheng2020structured3d}, which consists of 3,500 houses with 21,773 rooms designed by professional artists. 
A single panoramic image and 3D scene layout are provided in each room.
We parse the scene layout using oriented bounding boxes for common indoor room types like the bedroom, kitchen, living room, study, and bathroom.
Then, we follow~\cite{wang2021sceneformer} to generate text prompts describing the scene layout. The filtered dataset for training and evaluation consists of 4,961 bedrooms, 1,848 kitchens, 3,039 living rooms, 698 studies, and 1500 bathrooms. For each room type, we use $80\%$ of rooms for training and the remaining for testing. 
Following DiffuScene~\cite{tang2023diffuscene}, we further qualitatively evaluate our layout generation on 3D-FRONT dataset~\cite{fu20213dfront}.

\noindent \textbf{Metrics:} Follow previous work~\cite{paschalidou2021atiss, tang2023diffuscene}, Frechet Inception Distance (FID)~\cite{heusel2017gans} and Kernel inception distance (KID)~\cite{bińkowski2021demystifying} are used to measure the plausibility and diversity of 1,000 synthesized scene layouts. We choose FID, CLIP Score (CS)~\cite{radford2021learning}, and Inception Score (IS)~\cite{salimans2016improved} to measure the image quality of generated panoramas. 
To compare the quality of 3D room models, we follow Text2Room~\cite{hollein2023text2room} to render images of the 3D room model and measure the CLIP Score (CS) and Inception Score (IS). We also conduct a user study and ask $61$ users to score Perceptual Quality (PQ) and 3D Structure Completeness (3DS) of the final room mesh on scores ranging from 1 to 5.

\noindent More details about data preprocessing, experimental settings, and baseline implementations can be found in supplementary file Sec.4 and Sec.5.

\subsection{Comparison with Previous Methods}
\label{sec:exp_room_gen}

\subsubsection{Qualitative Comparison}
\myFigRef{fig:exp_qual_comp} shows some results generated by different methods. The first row shows a textured 3D room model, and the second row shows some perspective renderings from the room model. As we can see, Text2Light~\cite{chen2022text2light} fails to ensure the loop consistency of the generated panorama, which leads to distorted geometry and unreasonable room model. 
Both MVDiffusion~\cite{tang2023mvdiffusion} and Text2Room~\cite{hollein2023text2room} can generate vivid local images as demonstrated by the perspective renderings in the second row. But they fail to capture the global room layout. These two methods often repeat a dominating object, e.g. the cabinet in the bedroom appears multiple times at different places and violate the room layout constraint. In comparison, our method does not suffer from these problems and generates high-quality results. More examples are provided in the Fig.12 in supplementary file.

\begin{figure*}[t]
    \vspace{-0.3in}
    \centering
    \includegraphics[width=0.95\linewidth]{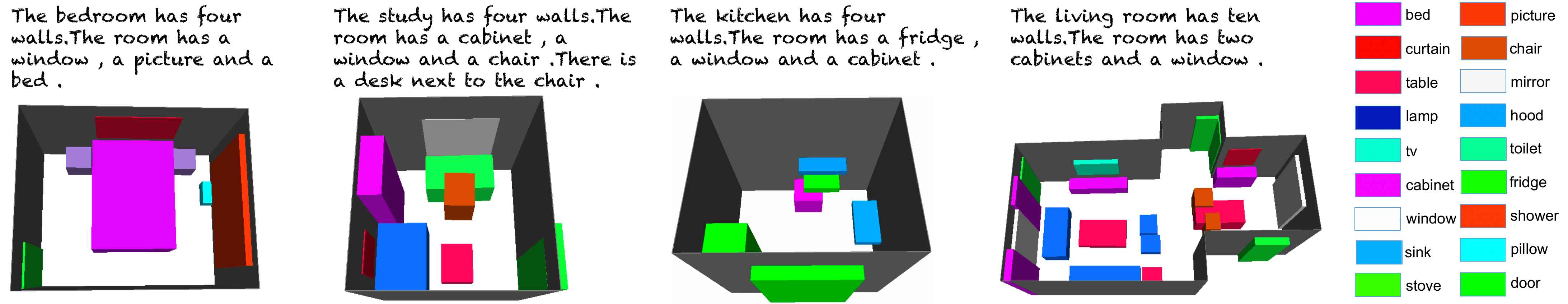}
    \vspace{-0.15in}
    \caption{Text-conditioned layout generation on Structured3D. Given the text prompt, our method synthesizes a plausible scene layout that matches the description. The generated layout is represented using different colors to indicate various object categories, such as blue for the sofa and brown for the chair. More results and semantic labels are provided in Fig.10 in supplementary file.
    }
    \label{fig:exp_layout_gen_results}
    \vspace{-0.1in}
\end{figure*}

\subsubsection{Layout Generation} 
\begin{table*}[t]
	\centering
        \caption{Quantitative Comparison of layout generation on 3D-FRONT. Note that DiffuScene-w-SC uses an additional network to learn a Shape Code for each furniture, facilitating the evaluation process to retrieve a more accurate CAD model for each furniture. Nevertheless, our method outperforms others in the common settings, where only the generated semantic class and size are used for retrieval. }
    \vspace{-0.1in}
	\resizebox*{0.8 \textwidth}{!}{
		\begin{tabular}{
				>{\centering\arraybackslash}p{4.8cm} 
                >{\centering\arraybackslash}p{4.8cm} 
				>{\centering\arraybackslash}p{0.9cm} 
				>{\centering\arraybackslash}p{0.9cm} 
				>{\centering\arraybackslash}p{0.9cm} 
				>{\centering\arraybackslash}p{0.9cm} 
				>{\centering\arraybackslash}p{0.9cm} 
				>{\centering\arraybackslash}p{0.9cm} 
			}
			\toprule
			\multirow{2}{*}{Method} & \multirow{2}{*}{Retrieval from} & \multicolumn{3}{c}{Livingroom} & \multicolumn{3}{c}{Diningroom}\\
			\cmidrule(lr){3-5} \cmidrule(lr){6-8} &
			&FID $\downarrow$ & KID $\downarrow$ & SCA &FID $\downarrow$ & KID $\downarrow$ & SCA  \\
			\midrule
              DiffuScene-w-SC~\cite{tang2023diffuscene} & Shape Code & \textbf{35.27} & \textbf{0.64} & \textbf{54.69} & \textbf{32.87} & \textbf{0.57} & \textbf{51.67} \\
		   \midrule
    	   ATISS~\cite{paschalidou2021atiss} & Semantic Bounding Box & 40.45 & 4.57 & 63.48 & 36.61 & 1.90 & 55.44\\
              DiffuScene-wo-SC~\cite{tang2023diffuscene} & Semantic Bounding Box  & 38.55 & \textbf{1.33} & 63.54 & 36.47 & 1.8 & 57.04 \\
              Ours & Semantic Bounding Box  & \textbf{36.0} & 1.4 & \textbf{56.42} & \textbf{34.78} &  \textbf{1.3} & \textbf{54.37} \\
			\bottomrule
		\end{tabular}
	}
	\label{tabel: comparision_layout_3dfront}
 \vspace{-0.1in}
\end{table*}

\noindent \myFigRef{fig:exp_layout_gen_results} verifies that our layout generation results are plausible and can offer reliable 3D scene layout constraints for the following appearance generation stage. As shown in ~\myFigRef{fig:exp_layout_gen_results}, our text-conditioned layout generation module can synthesize natural and diverse typical indoor scenes. The size and spatial location of the furniture are reasonable, and the relative positions between the furniture pieces are accurately recovered. Additional objects not described in the text are automatically generated according to the scene prior. 

\noindent 
Table~\ref{tabel: comparision_layout_3dfront}  provides a quantitative evaluation against state-of-the-art scene synthesis methods including ATISS~\cite{paschalidou2021atiss} and DiffuScene~\cite{tang2023diffuscene} on the 3D-FRONT. Following these methods, we rendered the generated scenes into $256\times256$ top-down orthographic images to compute the FID, KID, and Scene Classification Accuracy (SCA) scores. To facilitate this computation, ATISS, DiffuScene-wo-SC(without shape code), and our method retrieve the most similar CAD model in the 3D-FUTURE~\cite{fu20213d} for each object based on generated semantic class and sizes. DiffuScene-w-SC uses an additional network to learn a shape code for each furniture to choose a better 3D mesh model. 
Note that the SCA score is better when it is closer to 50$\%$. We have excluded walls, doors, and windows from our scene code representation to ensure a fair comparison. Table~\ref{tabel: comparision_layout_3dfront} shows our method achieves results superior to that of ATISS and DiffuScene-wo-SC, indicating that our approach is capable of producing more realistic and natural layouts of indoor scenes.

\subsubsection{Panorama Generation}
\begin{figure*}[t]
    \vspace{-0.3in}
    \centering
    \includegraphics[width=0.9\linewidth]{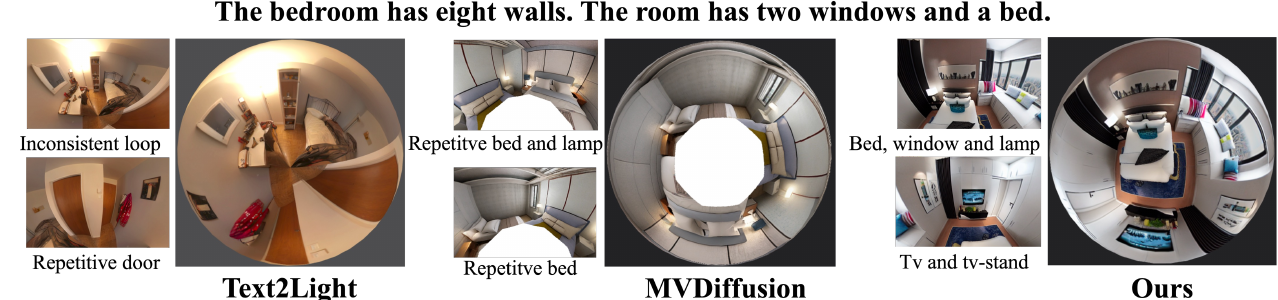}
    \vspace{-0.1in}
    \caption{Qualitative comparison for panorama generation. More results are available in the  Appendix. 
    }
    \label{fig:exp_pano_gen_results}
    \vspace{-0.1in}
\end{figure*}

\noindent \myFigRef{fig:exp_pano_gen_results} qualitatively evaluates our generated panoramic images, the image is visualized in a panoramic image viewer to facilitate the user to check the global content. The left side of each column is two zoom-in views, and the right side is the fisheye view. Text2Light~\cite{chen2022text2light} suffers from serious inconsistency on the borders of the generated panorama. It also shows a lot of unexpected objects in the image. MVDiffusion~\cite{tang2023mvdiffusion} suffers from repetitive furniture and fails to synthesize reasonable content for the target room type. In contrast, our method obtains a plausible layout and vivid panorama from the given text prompt. 

\begin{table*}[t]
        \vspace{-0.0in}
	\centering
        \caption{Quantitative Comparison of panorama and mesh generation.}
    \vspace{-0.1in}
	\resizebox*{0.8 \textwidth}{!}{
		\begin{tabular}{
				>{\centering\arraybackslash}p{4.8cm} 
				>{\centering\arraybackslash}p{0.9cm} 
				>{\centering\arraybackslash}p{0.9cm} 
				>{\centering\arraybackslash}p{0.9cm} 
				>{\centering\arraybackslash}p{0.9cm} 
				>{\centering\arraybackslash}p{0.9cm} 
				>{\centering\arraybackslash}p{0.9cm} 
				>{\centering\arraybackslash}p{0.9cm} 
			}
			\toprule
			\multirow{2}{*}{Method} & \multicolumn{3}{c}{Panorama Metrics} & \multicolumn{2}{c}{2D Rendering Metrics} & \multicolumn{2}{c}{3D Mesh User Study}\\
			\cmidrule(lr){2-4} \cmidrule(lr){5-6} \cmidrule(lr){7-8}
			&FID $\downarrow$ & CS $\uparrow$ & IS $\uparrow$  &CS $\uparrow$ & IS $\uparrow$ & PQ$\uparrow$ &3DS $\uparrow$ \\
			\midrule
    	   Text2Light~\cite{chen2022text2light} & 56.22 & 21.45 & \textbf{4.198} & - & - & 2.732 & 2.747\\
              MVDiffusion~\cite{tang2023mvdiffusion} &34.76 & \textbf{23.93} &3.21 & - & - & 3.27 & 3.437 \\
              Text2Room~\cite{hollein2023text2room} & - & - & - & 25.90 & 2.90 & 2.487 & 2.588 \\
              Ours & \textbf{21.02} & 22.19 & 3.56 &  \textbf{25.97} & \textbf{3.14} & \textbf{3.89} & \textbf{3.746}\\
			\bottomrule
		\end{tabular}
	}
	\label{tabel: comparision_mesh}
\end{table*}

\begin{figure}
    \vspace{-0.1in}
    \centering
    \includegraphics[width=0.9\linewidth]{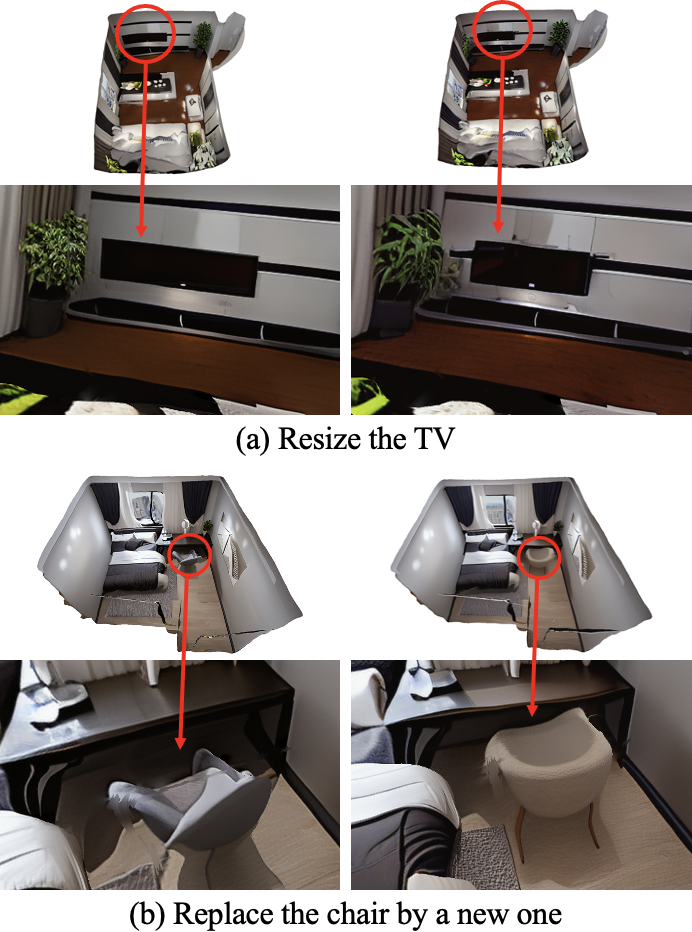}
    \vspace{-0.15in}
    \caption{Editing examples. (a) resize the TV, (b) replace the chair with a new one.  }
    \label{fig:exp_scene_edit}
    \vspace{-0.2in}
\end{figure}

\noindent Table~\ref{tabel: comparision_mesh} provides quantitative evaluations. We follow MVDiffusion~\cite{tang2023mvdiffusion} to crop perspective images from the generated panoramas on the test split and evaluate the FID, CS, and IS scores on the cropped multi-view images. 
In the left part of Table~\ref{tabel: comparision_mesh}, our method achieves the best score in FID, which indicates that our method can better capture the room appearance because of its faithful recovery of the room layout.
However, our score on CS is slightly lower than MVDiffusion, which seems insensitive to the number of objects and cannot reflect the room layout quality. The IS score depends on the semantic diversity of the cropped images as captured by an image classifier. Text2Light has the best IS score, since the generations contain unexpected objects.


\noindent 
In Fig.8 of the supplementary file, we also study the
performance of our panorama generation module with and without loop-consistent sampling mechanism, the ablation indicates the loop-consistent sampling helps the generated panorama obtain better texture consistency.

\subsubsection{3D Room Generation}
We then compare the 3D room models in terms of their rendered images. Because of the expensive running time of Text2Room~\cite{hollein2023text2room}, we only test on 12 examples for this comparison. 
In this comparison, we further skip Text2light and MVDiffusion since we have compared them on panoramas.
As the room layout is better captured with a large FOV, we render 60 perspective images of each scene with a $140^\circ$ FOV and evaluate their CS and IS scores respectively. 
The results of this comparison are shown in the middle of Table~\ref{tabel: comparision_mesh}. Our method obtains better scores on both metrics than Text2Room. 

\noindent We further evaluate the quality of the textured 3D mesh model by user studies. 
The results of the user study are shown on the right of  Table~\ref{tabel: comparision_mesh}. Users prefer our method over others, for its clear room layout structure and furniture arrangement.


\subsection{Interactive Scene Editing}
\label{sec:exp_scene_editing}
We demonstrate the scene editing capability of our method in ~\myFigRef{fig:exp_scene_edit}. In this case, we resize the TV and replace the chair in the generated results. Fig.~\ref{fig:teaser_a} (b) shows examples of replacing the TV and TV stand. Our method can keep the visual appearance of the moved/resized objects unchanged after editing. More examples can be found in the appendix.

 \section{Conclusion}
\label{sec:conclusion}
We present \textbf{Ctrl-Room}, a flexible method to achieve structurally plausible and editable 3D indoor scene generation. It consists of two stages, the layout generation stage and the appearance generation stage. In the layout generation stage, we design a scene code to parameterize the scene layout and learn a text-conditioned diffusion model for text-driven layout generation. In the appearance generation stage, we fine-tune a ControlNet model to generate a vivid panorama image of the room with the guidance of the layout. Finally, a high-quality 3D room with a structurally plausible layout and realistic textures can be generated via the layout-guided panoramic NeRF. We conduct extensive experiments to demonstrate that \textbf{Ctrl-Room} outperforms existing methods for 3D indoor scene generation both qualitatively and quantitatively, and supports interactive 3D scene editing.

\section{Limitation}
\label{sec:limitation}

There are still some limitations of Ctrl-Room. Firstly, we only support single-room generation, thus we cannot produce large-scale indoor scenes with multiple rooms. 
A promising direction is to learn a text-driven diffusion model to produce more consistent RGB-D panorama images cross multiple rooms under the scene layout constraints. Secondly, as we explore injecting 3D scene information into pretrained 2D models, thus we rely on 3D labeled scene dataset to drive the learning and fine-tuning process. Leveraging scene datasets with only 2D labels to learn 3D priors is also a promising direction. Thirdly, the generated 3D model still contains artifacts and incomplete structures in invisible areas because of the occlusion and poor performance of the panoramic depth estimator. We leave the aforementioned limitations for our future efforts.

\clearpage
\setcounter{page}{1}
\maketitlesupplementary

\noindent In the supplementary file, we first present more details about our scene code diffusion model in \mySecRef{sec:append:scene_code_diffusion}, then we elaborate the layout-guided PeRF module and mask-guided editing method in \mySecRef{sec:append:layout_guided_perf} and \mySecRef{sec:append:mask_guided_editing}, respectively. Next, we provide our dataset pre-processing, text prompt generation, and implementation details in \mySecRef{sec:append:dataset} and \mySecRef{sec:append:impl} respectively. Additional experiment results are also illustrated, including panorama generation comparisons in \mySecRef{sec:append:additional_pano_gen_results}, room layout generations and room mesh comparisons in \mySecRef{sec:append:additional_qual_results} and user studies in \mySecRef{sec:append:user_study}. Furthermore, we demonstrate that our scene code diffusion model can be trained with free-style text prompts in \mySecRef{sec:append:free_style_prompt}.

\section{Scene Code Denoising Network}
\label{sec:append:scene_code_diffusion}

\begin{figure}
    \centering
    \includegraphics[width=\linewidth]{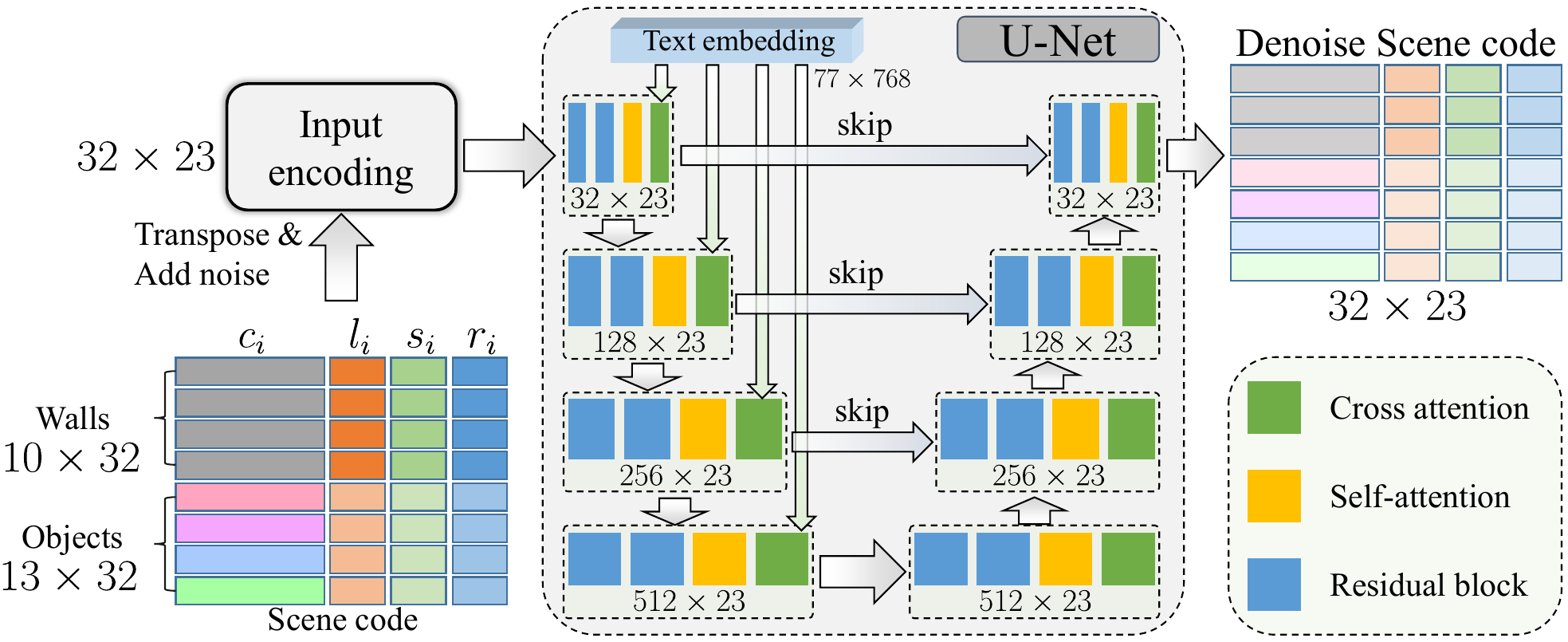}
    \caption{The detailed structure of the scene code denoising network. We here take the bedroom for example to demonstrate the dataflow of the scene code denoiser. The scene code tensor $\mathbf{x_0} \in \mathbb{R}^{N\times D}$, where $N=23, D=32$.}
    \label{fig:supp_unet}
\end{figure}

In the Layout Generation Stage, we use a holistic scene code to parameterize the indoor scene and design a diffusion model to learn its distribution. Specifically, given a 3D scene $\mathcal{S}$ with $N$ objects, we represent the scene layout as a holistic scene code $\mathbf{x_0}=\{ \mathbf{o_i} \}^{N}_{i=1}$. We encode each object $o_i$ as a node with various attributes, i.e., center location $l_i \in \mathbb R^3$, size $s_i \in \mathbb R^3$, orientation $r_i \in \mathbb R$, class label $c_i \in \mathbb R^{C}$. Each node is characterized by the concatenation of these attributes as $\mathbf o_i = [c_i, l_i, s_i, r_i]$. As shown in \myFigRef{fig:supp_unet}, our scene code denoising network of the layout diffusion model is built upon IDDPM~\cite{nichol2021improved}. The whole architecture of the layout diffusion model is similar to IDDPM, while we replace the upsample and downsample blocks with 1D-convolution network in the U-Net, and insert attention blocks after each residual block to capture both the global context among objects and the semantic context from the input text prompt. The input encoding head processes different encoding of the node attributes, e.g., semantic class labels, box centroid, and box orientation. After adding noise, the input encoding is fed into the U-Net to obtain a denoised scene code. 
The training objectives includes the denoising objective $\mathcal{L_{\rm denoise}}$ and a regularization term $\mathcal L_{\rm physical}$ to penalize the penetration among objects and walls as follows,
\begin{eqnarray}
    \mathcal{L} & = & \mathcal{L_{\rm denoise}} + \mathcal L_{\rm physical}, \\
    \mathcal{L_{\rm denoise}} & = & \mathbf{E}_{\mathbf{x_0}, t, y, \mathbf{\epsilon}} {\left \| \mathbf{\epsilon} - \epsilon_{\theta}(x_t,t, y)  \right \|^2}, \\
    \mathcal L_{\rm physical} & = & \sum_{t=1}^{T} w_t \ast (\mathcal L_{\rm w-o} + \mathcal L_{\rm o-o}). 
\label{eq:append_layout_diffusion_physical_objective}
\end{eqnarray}
where $\epsilon_{\theta}$ is the noise estimator which aims to find the noise $\mathbf \epsilon $ added into the input $x_0$. Here, $y$ is the text embedding of the input text prompts. The hyperparamter $w_t$ is set to $\mathbf{\bar {\alpha}_t}* 0.1$.
$\mathcal L_{\rm w-o}$ is the physical violation loss between walls and objects. It is defined as follows,

\begin{equation}
\begin{split}
    \mathcal L_{\rm \mathbf{w-o}} &= \sum_{i=1}^{K_{\rm wall}} \sum_{j=1}^{K_{\rm object}} \sum_{p=1}^{8}  Relu[-(a_i x_{jp}+b_i y_{jp}+c_i z_{jp}+d_i)] \\
    &*  \mathds{1} (\prod_{\mathbf w_i} (x_{jp}, y_{jp}, z_{jp}) \, {\rm in} \, \mathbf w_i).
\end{split}
\end{equation}
Here, $(a_i, b_i, c_i)$ is the normal vector of wall $\mathbf w_i$ that points towards the room center. $\prod_{\mathbf w_i}$ is the operator projecting a
point onto the plane defined by $\mathbf w_i$. The plane equation of i-th
wall is $a_i x + b_i y + c_i z + d = 0$ and $\mathds{1} (\prod_{\mathbf w_i} (x_{jp}, y_{jp}, z_{jp}) \, {\rm in} \, \mathbf w_i)$
indicates whether the projection of bounding box vertices $(x_{jp}, y_{jp}, z_{jp})$ of j-th object is inside $\mathbf w_i$.
We skip some objects such as windows and doors since they can intersect with walls. We adopt the 3D IoU loss $\mathcal L_{\rm o-o}$ in DiffuScene~\cite{tang2023diffuscene} as follows,
\begin{equation}
        \mathcal L_{\rm \mathbf{o-o}}  =  \sum_{\mathbf o_i, \mathbf o_j}^{K_{\rm object}} \mathbf {IoU(o_i, o_j)}.
\end{equation}

\noindent During the forward phase, as in IDDPM, we iteratively perform the denoising process and generate a scene code from a partial scene textual description.

\begin{figure*}[!t]
    \vspace{-0.1in}
    \centering
    \includegraphics[width=0.9\linewidth]{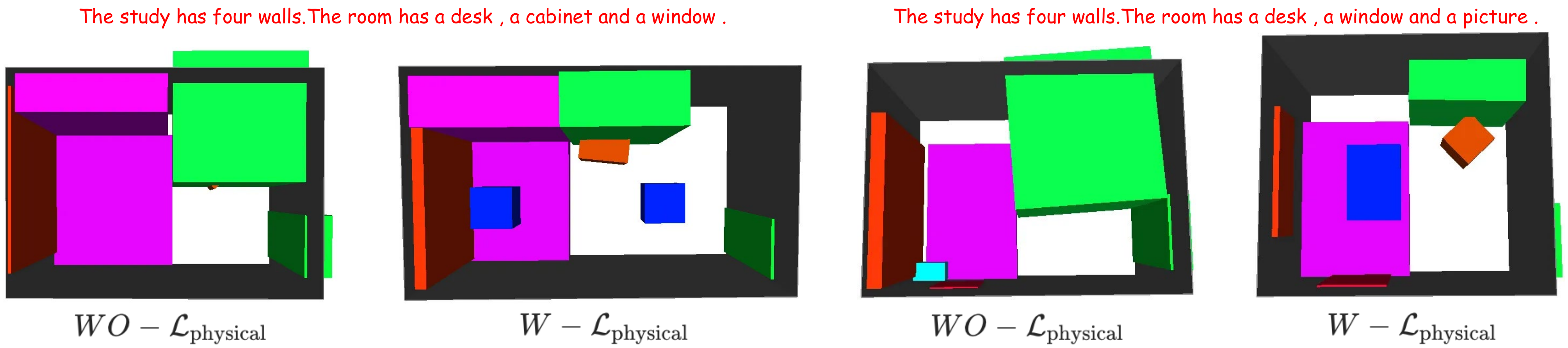}
    \caption{Ablation study of the physical violation loss. Two text prompts of study are used for layout generation using our diffusion model trained without $\mathcal{L_{\rm physical}}$ (left) and with $\mathcal{L_{\rm physical}}$ (right), respectively. As a result, in the left sample, diffusion model without $\mathcal{L_{\rm physical}}$ generates a green desk that penetrates the wall. In the right sample, this phenomenon is alleviated and regulated after using the physical violation loss. Note that the sampling results of these two versions of diffusion models are slightly different since the denoise distribution is different even given the same text prompt.  }
    \label{fig:ablation_physical_loss}
    \vspace{-0.1in}
\end{figure*}

\noindent We further investigate how the physical regularization term impacts the final 3D scene layout. In~\myFigRef{fig:ablation_physical_loss}, we use two text prompts for layout generation our layout diffusion model trained with and without our physical regularization term, respectively. As can be seen, the diffusion model trained with the physical violation loss can effectively reduce the occurrence of furniture penetrating walls, and also help to regulate the orientation of the sampled furniture, resulting in more reasonable layouts than the model without the physical regularization term.

\begin{figure}
    \centering
    \includegraphics[width=\linewidth]{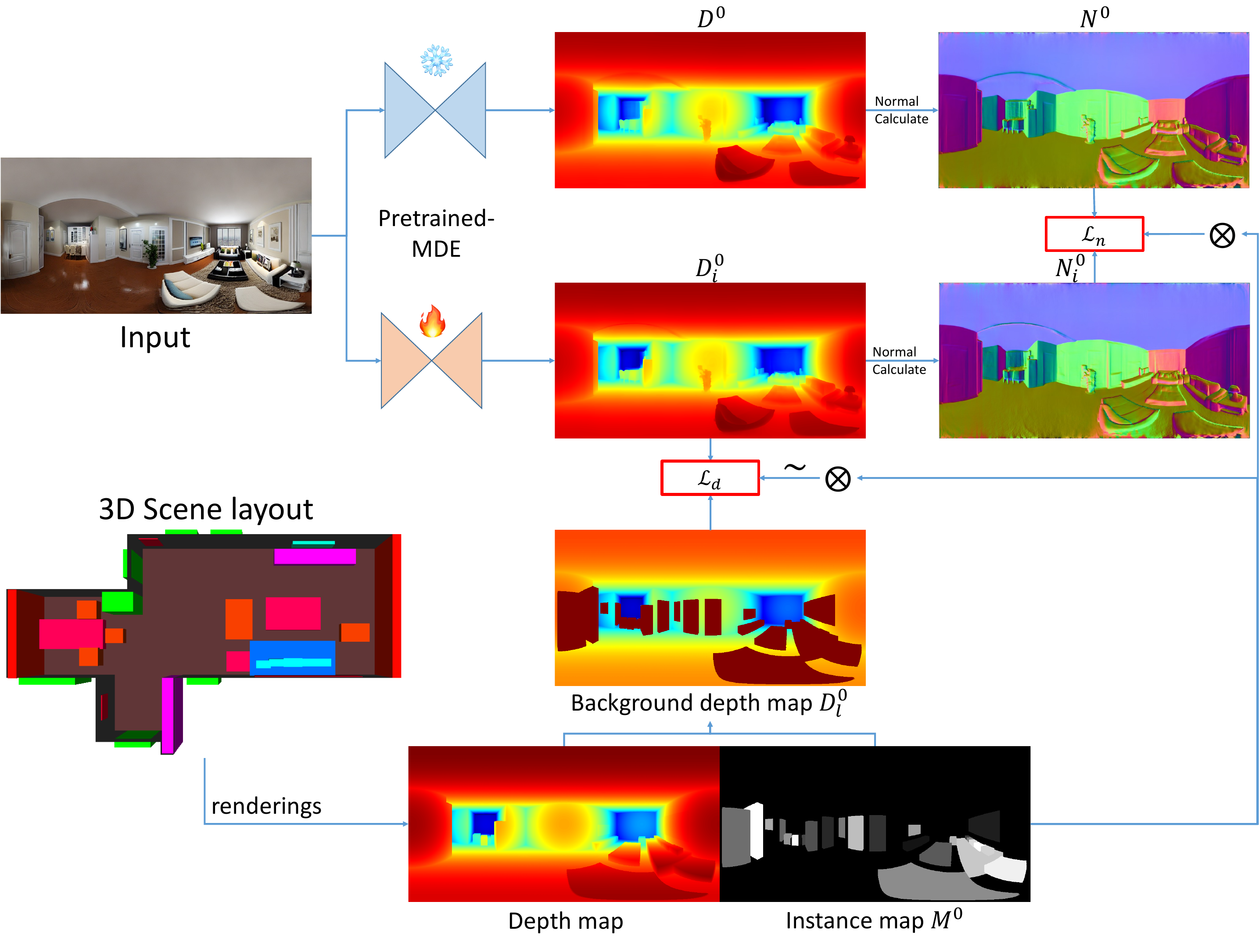}
    \caption{Our Layout-guided depth estimation. To align the estimated depth map with the 3D scene layout, we render depth map and instance map from the 3D scene layout at current view. Then we take the background depth (wall, ceiling, floor) as reference to align the depth prediction by optimizing the pretrained MDE. Avoiding degrade the MDE's performance at object surface, we ensure the normal consistency of each object during the optimization process.  }
    \label{fig:panoramic_geometry_align}
    \vspace{-0.2in}
\end{figure}

\section{Layout-Guided Panoramic NeRF}
\label{sec:append:layout_guided_perf}
Since the generated panorama is a partial observation of the room subject to occlusions, lifting the single view into a 3D room becomes a complex problem. Here we adopt the Warp and Inpainting scheme to complete the 3D room progressively. 
After generating the panorama $I^0$, we recover its depth map $D^0$ using the state-of-the-art monocular depth estimator (MDE) ~\cite{yun2023egformer}. However, problems such as scale ambiguity and large occlusions may lead to incomplete 3D room reconstruction. Additionally, ensuring consistency in inpainted panoramas at new viewpoints poses another challenge. 
Fortunately, the scene layout generated in the first stage offers crucial geometric and semantic guidance, which can help correct biased depth predictions and guide the inpainting process to generate novel view panoramas. 
In this paper, we employ PeRF~\cite{wang2023perf} as the 3D room model and progressively generate novel viewpoint panoramas with layout guidance to reconstruct the PeRF model.

\begin{figure*}[ht]
    \vspace{-0.1in}
    \centering
    \includegraphics[width=0.9\linewidth]{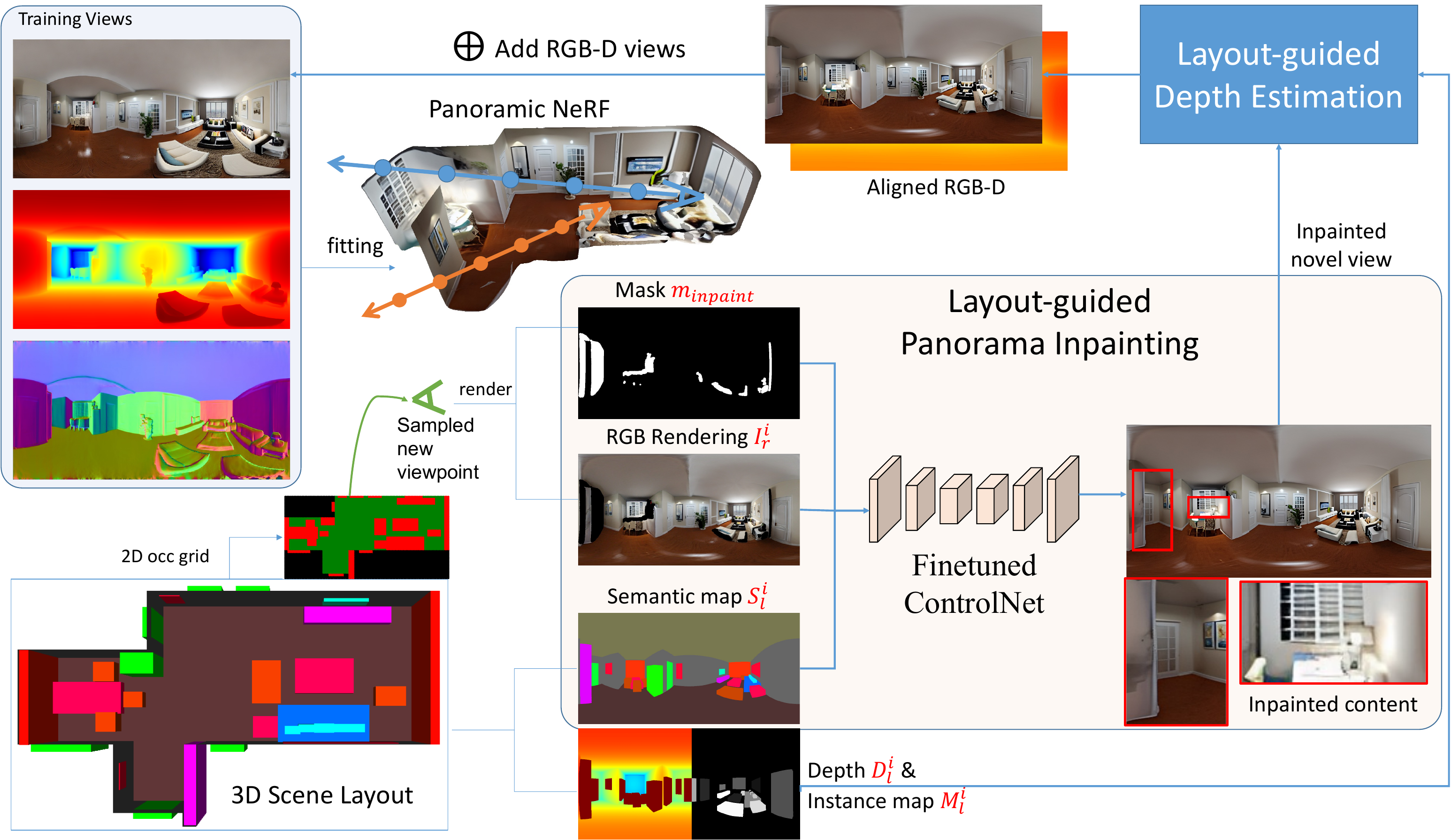}
    \caption{The Layout-guided PeRF takes the input panorama, aligned depth map and normal map as initialization. Then a progressive inpainting module is introduced to generate consistent panoramic images at the sampled novel views. The progressive inpainting module consists of the layout-guided panorama inpainting and the layout-guided depth estimation module. The final RGB-D panoramic pairs are included as training views to finetune PeRF~\cite{wang2023perf}. }
    \label{fig:panoramic_view_completion}
    \vspace{-0.2in}
\end{figure*}

\noindent \textbf{Layout-Guided Depth Estimation.} 

To align the estimated panoramic depth map with the scene layout, a naive approach would be to directly compute scale and bias coefficients for the initial depth map $D^0$. However, as the scene layout consists of object bounding boxes and can not provide pixel-level perfect depth supervision, this method may lead to degraded depth predictions. To address this problem, we propose the panoramic geometry alignment module as depicted in \myFigRef{fig:panoramic_geometry_align}. We use the instance labels of furniture items to exclude the rendered depths within the furniture areas, retaining only the background depth map (e.g., wall, ceiling, floor) denoted as $D^0_l$. We incorporate a consistency loss $\mathcal{L}_{\rm align}$ to optimize the pre-trained monocular depth estimator (MDE). This consistency loss is formulated as follows,
\begin{eqnarray}
    \mathcal{L}_{\rm align} & = & \mathcal{L_{\rm d}} * w_d + \mathcal L_{\rm n} * w_n, \\
    \mathcal{L_{\rm d}} & = & L_1 (D^0_i, D^0_l) * (\sim M^0) , \\
    \mathcal L_{\rm n} & = & \left | N^0 - N^0_i \right | * M^0. 
\label{eq:append_geometry_align_loss}
\end{eqnarray}
where $\mathcal L_{\rm d}$ represents the smooth L1 loss of depth, $\mathcal L_{\rm n}$ denotes the absolute loss of normal, and $w_d$, $w_n$ are weighting coefficients. $D^0_i$ stands for the predicted depth of the MDE at i-th iteration. $M^0$ is the instance map denoting the foreground, such that we only correct the predicted depth in the background region while preserving normals in the furniture regions. After alignment, we obtain the optimal depth map $D^{*}$ and normal map $N^{*}$.

\noindent \textbf{Layout-Guided Novel View Generation.}
PeRF~\cite{wang2023perf} trains a panoramic neural radiance field using a single panorama. To render novel view panoramas, it employs the Stable Diffusion model~\cite{rombach2022high} to inpaint 60 perspective images and stitch them into a panorama. Although it produces consistent renderings at nearby viewpoints, it struggles to synthesize viewpoints that are far apart and involve large unoccluded regions. To address this limitation, we rely on the scene layout to guide the panorama inpainting to maintain cross-view consistency. 

\begin{figure}
    \centering
    \includegraphics[width=\linewidth]{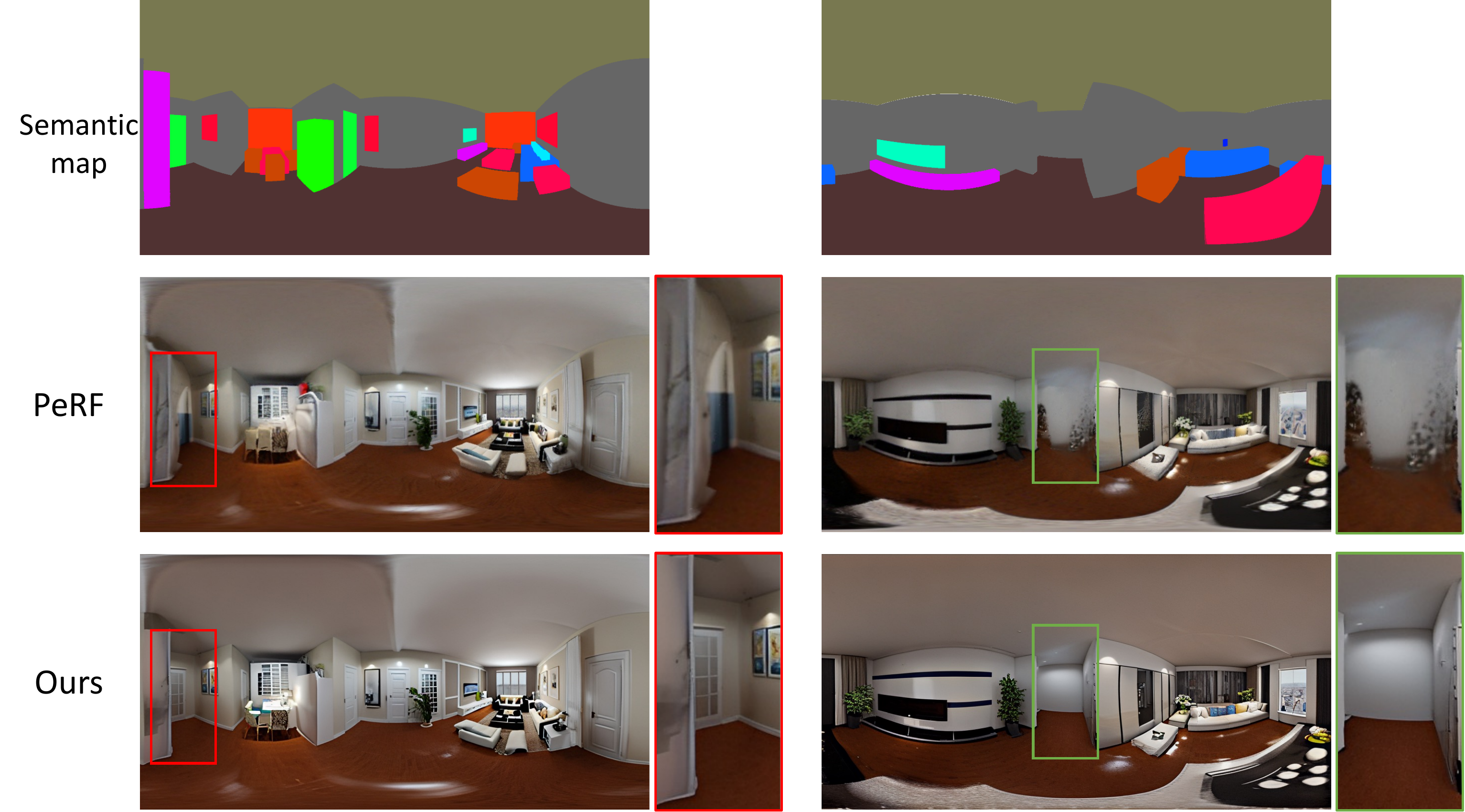}
    \caption{After the layout-guided panorama inpainting, Our generated panoramas at novel viewpoints adhere to the semantic layout and seamlessly integrate with the visible regions, while PeRF~\cite{wang2023perf} fails to synthesize plausible content at those large-size occlusion areas. }
    \label{fig:panoramic_view_completion_demo}
\end{figure}

\noindent As illustrated in \myFigRef{fig:panoramic_view_completion}, we initialize the scene NeRF with the panorama $I^0$, the aligned depth map $D^*$ and normal map $N^*$. We sample new viewpoints in the green area of the occupancy grid that do not conflict with the initial furniture arrangement. At the i-th novel view, we render semantic map $S^i_l$, depth map $D^i_l$ and instance map $M^i_l$ from the scene layout, these are then combined with the panoramic rendering $I^i_r$ and inpainting mask $\mathbf m_{\rm inpaint}$ obtained from the NeRF and fed to the layout-guided panorama inpainting module to generate the novel view panorama. Using our fine-tuned ControlNet, it achieves training-free panoramic inpainting, which replaces pixels outside the inpainting mask $\mathbf m_{\rm inpaint}$ with $I^i_r$ and fill $\mathbf m_{\rm inpaint}$ based on the semantic map $S^i_l$. 
As demonstrated in \myFigRef{fig:panoramic_view_completion_demo}, our resulting RGB panorama adheres to the semantic layout and seamlessly integrates with the visible regions, while that of PeRF~\cite{wang2023perf} fails to generate reasonable content in the large occlusion areas.
Subsequently, after generating the novel view panorama, we apply the layout-guided depth estimation and include it as training views for PeRF following their framework~\cite{wang2023perf}.

\begin{figure*}[t]
\vspace{-0.1in}
\centering
\includegraphics[width=0.95\textwidth]{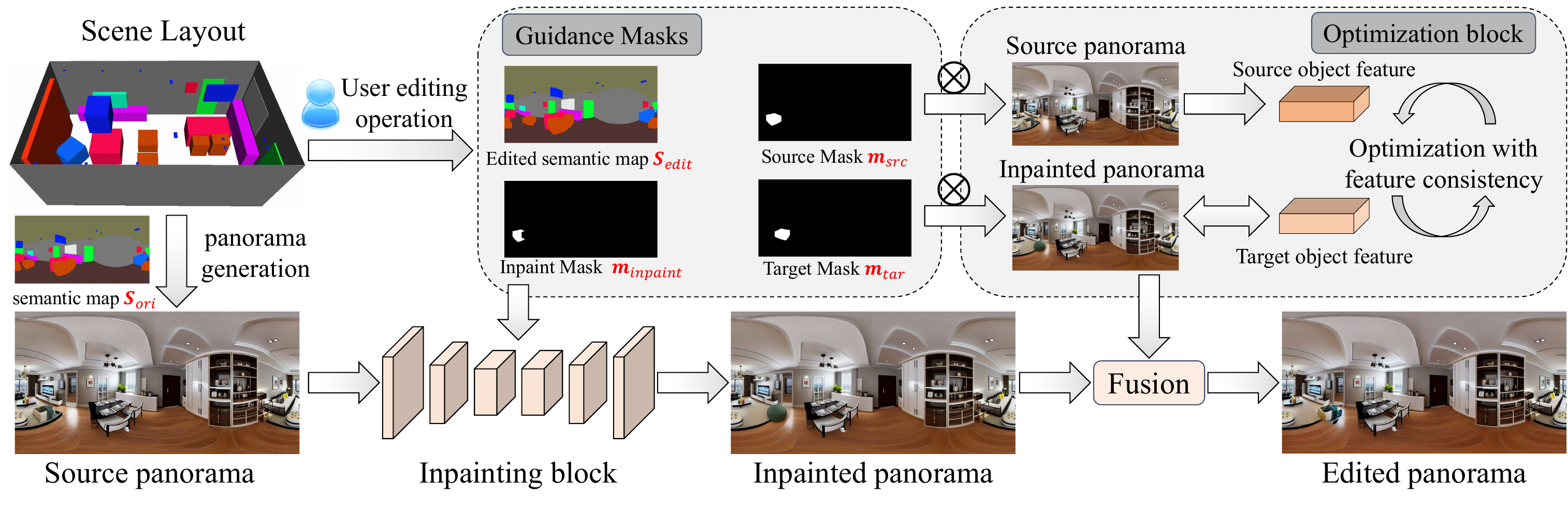}
\vspace{-0.2in}
\caption{\textbf{Mask-guided Editing.} After editing the scene bounding box, we derive guidance masks from the changes in the semantic layout panoramas. We fill in unoccluded regions and optimize the DIFT~\cite{tang2023emergent} features to keep the identity of moved objects unchanged.
}
\label{fig:pano_edit_module}
\vspace{-0.1in}
\end{figure*}

\section{Mask-Guided Editing}
\label{sec:append:mask_guided_editing}
To achieve consistent and seamless 3D scene editing, it should achieve two goals, i.e. altering the content according to the user's input, and maintaining appearance consistency for scene objects. 
We propose a mask-guided image editing as illustrated in \myFigRef{fig:pano_edit_module}, where a chair's position is moved. In the following, we will explain our method with this example.
We denote the semantic panorama from the edited scene as $S_{\rm edited}$, then we derive the guidance masks based on its difference from the original one $S_{\rm ori}$. The source mask $\mathbf m_{\rm src}$ shows the position of the original chair, and the target mask $\mathbf m_{\rm tar}$ indicates the location of the moved chair, and the inpainting mask $\mathbf m_{\rm inpaint} = \{m | m \in \mathbf m_{\rm src} \; {\rm and} \; m \notin \mathbf m_{\rm tar} \}$ is the unoccluded region.
Given these guidance masks, our method includes two steps: the inpainting step and the optimization step. We first fill in the inpaint area by feeding the inpaint mask $\mathbf m_{\rm inpaint}$ and edited semantic panorama $S_{\rm edited}$ to the inpainting step. Then, in our optimization step, we optimize the DIFT~\cite{tang2023emergent} feature to maintain the visual consistency of relocated objects.

\noindent
\textbf{Inpainting Step.} 
Denoting the original image as $\mathbf{x^{{\rm ori}}_{0}}$, we replace pixels outside the inpainting mask $\mathbf m_{\rm inpaint}$ with  $\mathbf{x^{{\rm ori}}_{t}}$ during the diffusion process. This simple strategy keeps the outside region unchanged. 
At each reverse diffusion step, we compute:
\begin{small}
\begin{eqnarray}
\label{eq:edit_inpaint_algo_1}
\mathbf{x_t^{\rm ori}} & \sim & \mathcal{N}(\sqrt{\bar{\alpha_t}}\mathbf{x_0^{\rm ori}} (1-\bar{\alpha_t}\mathbf{I})), \\
\label{eq:edit_inpaint_algo_2}
\mathbf{x_t^{{\rm new}}} & \sim & \mathcal{N}(\mu_\theta(x_t,t,y, S_{\rm edited}), \Sigma_{\theta}(x_t,t,y,S_{\rm edited})) \\
\label{eq:edit_inpaint_algo_3}
\hat{\mathbf x}_{t-1}^{\rm new} & = & \mathbf m_{\rm inpaint} \odot \mathbf{x_t^{{\rm new}}} + (1- \mathbf m_{\rm inpaint}) \odot \mathbf{x_t^{\rm ori}}
\end{eqnarray}
\end{small}
where $\mathbf{x^{{\rm ori}}_{t}}$ is obtained through propagating $\mathbf{x^{{\rm ori}}_{0}}$ in diffusion process, and $\mathbf{x_t^{{\rm new}}}$  is sampled from the fine-tuned ControlNet model, which takes the edited semantic layout panorama $S_{\rm edited}$ and text prompt $y$ as input.
As the propagated $\mathbf{x^{{\rm ori}}_{t}}$ is unaware of the new content $\mathbf{x_t^{{\rm new}}}$, this may result in distracting boundaries of the inpainted area. To better blend the new content $\mathbf{x}^{\rm new}_t$ and its surrounding background $\mathbf{x^{{\rm ori}}_{t}}$ in the inpainted area, we update the computation of $\hat{\mathbf x}_{t-1}^{\rm new}$ to,
\begin{equation}
\begin{split}
    \hat{\mathbf x}_{t-1}^{\rm new} &= \mathbf m_{\rm inpaint} \odot \mathbf{x_{t}^{{\rm new}}} \\
    & + (1- \mathbf m_{\rm inpaint}) \odot (\mathbf{x_{t}^{\rm ori}} \cdot \lambda_{\rm ori} + \mathbf{x_{t+1}^{\rm new}} \cdot \lambda_{\rm new} )
\end{split}
\label{eq:edit_inpaint_algo_4}
\end{equation}
where $\lambda_{\rm ori}$ and $\lambda_{\rm new}$ are hyper-parameters to adjust the weight for fusing the inpainted area and unchanged area. The final result of inpainting is $\hat{\mathbf x}^{\rm new}_0$.

\noindent
\textbf{Optimization Step.} 
When the user moves the position of a furniture item, we need to keep its appearance unchanged before and after the movement. 
The recent work, DIFT~\cite{tang2023emergent}, finds the learned features from the diffusion network allow for strong semantic correspondence.
Thus, we maintain the consistency between the original and moved furniture by requiring their latent features to be consistent. In particular, we extract latent features $F^{l}_t$ of the layer $l$ in the denoising U-Net network, at timestep $t$. Then we construct a loss function using the latent features from source area $\mathbf m_{\rm src}$ in source panorama $\mathbf{x^{{\rm ori}}_{0}}$ and target area $\mathbf m_{\rm tar}$ in inpainted panorama $\hat{\mathbf x}^{\rm new}_0$. 

For conciseness, we denote the target image $\hat{\mathbf x}^{\rm edit}_0$ initialized by $\hat{\mathbf x}^{\rm new}_0$.
We first propagate the original image $\mathbf{x_0^{\rm ori}}$ and $\hat{\mathbf x}^{\rm edit}_0$ to get $\mathbf{x_t^{\rm ori}}$ and $\hat{\mathbf x}^{\rm edit}_{t}$ at timestep $t$ by diffusion process, respectively. At each iteration, we use the same ControlNet model to denoise both $\mathbf{x_{t}^{\rm ori}}$ and $\hat{\mathbf x}^{\rm edit}_{t}$ and extract the latent features of them, denoted as $F^{\rm ori}_t$ and $F^{\rm edit}_t$, respectively. Based on the strong correspondence between the features, the source mask area $\mathbf m_{\rm src}$ and the target area $\mathbf m_{\rm tar}$ in $F^{\rm ori}_t$ and $F^{\rm edit}_t$ need to have high similarity. Here, we utilize the cosine embedding loss to measure the similarity, and define the optimization loss function as follows: 
6.
Here, ${\rm sg}$ is the stop gradient operator, the gradient will not be back-propagated for the term ${\rm sg}(F^{ori}_t\odot \mathbf m_{\rm src})$. Then we minimize the loss iteratively. At each iteration, $\hat{\mathbf x}^{\rm edit}_{t}$ is updated by taking one gradient descent step with a learning rate $\eta$ to minimize the loss $\mathcal{L_{\rm obj}}$ as,
\begin{equation}
    \hat{\mathbf x}^{k+1}_{t} = \hat{\mathbf x}^{k}_{t} - {\eta} \cdot \frac{\partial \mathcal{L_{\rm obj}}}{\partial \hat{\mathbf x}^{k}_{t}} 
    \label{eq:edit_opt_algo_2}
\end{equation}
After $M$ steps optimization, we apply the standard denoising process to get the final result $\hat{\mathbf x}^{\rm edit}_{0}$.

\section{Dataset}
\label{sec:append:dataset}

\begin{figure*}[!t]
    \centering
    \includegraphics[width=0.75\linewidth]{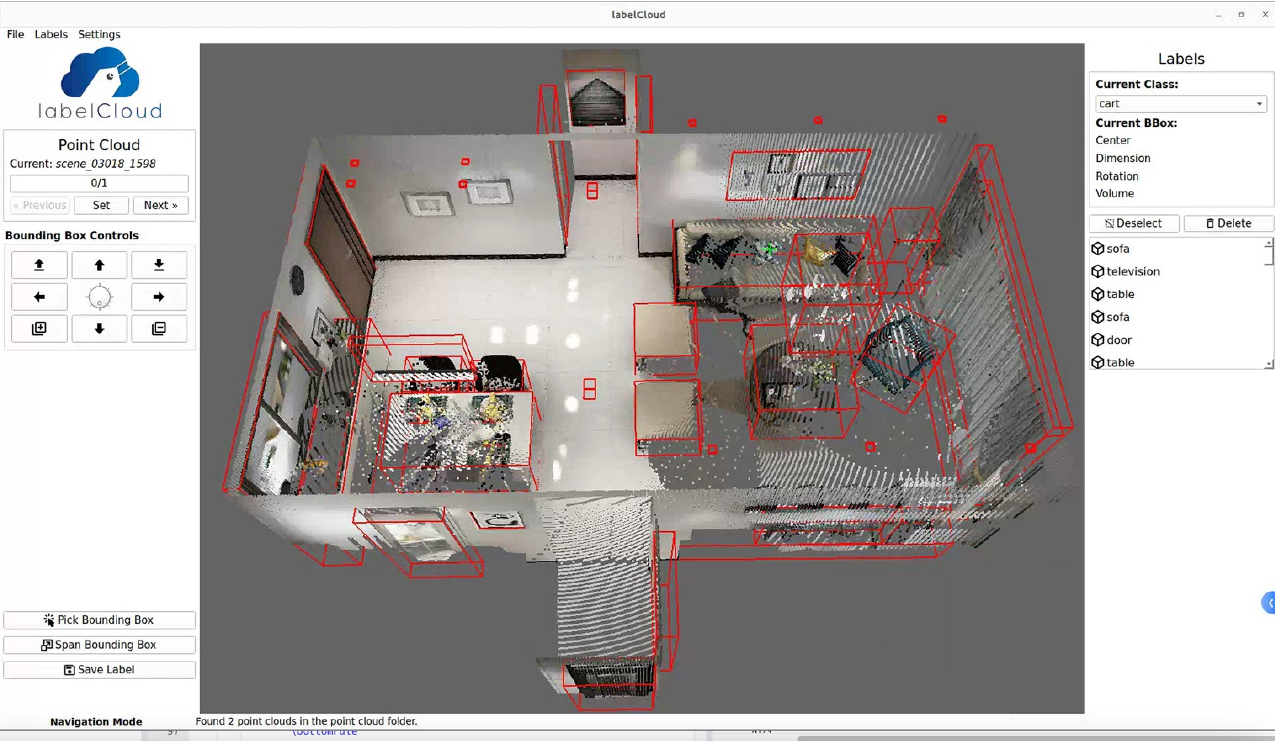}
    \caption{Example of object bounding box annotation.}
    \label{fig:append_dataset_annotation}
\end{figure*}

\noindent \textbf{Structured3D dataset preprocessing} Structured3D consists of $3,500$ houses with $21,773$ rooms, where each room is designed by professional designers with rich 3D structure annotations, including the room planes, lines, junctions, and orientated bounding box of most furniture, and photo-realistic 2D renderings of the room. In our work, we use the 3D orientated bounding boxes of furniture, 2D RGB panorama, and 3D lines and planes of each room. While the original dataset lacks semantic class labels for each furniture bounding box. The dataset preprocessing aims to produce clean ground truth data for our layout generation module and appearance generation module. 
\begin{itemize}
    \item \textbf{Orientated Object Bounding Box Annotation.} As the original dataset lacks semantic label for each orientated object bounding box, we first unproject the RGB panorama and depth map into a point cloud of the room, then manually annotate the object semantic class and add more accurate object bounding boxes based on the noisy annotation of the original version. As shown in \myFigRef{fig:append_dataset_annotation}, by using labelCloud~\cite{Sager_2022}, three data annotators worked for 1200 hours to annotate 5,064 bedrooms, 3,064 livingrooms, 2,289 kitchens, 698 studies, and 1,500 bathrooms, getting nearly 150K accurate orientated 3D bounding boxes across 25 object categories. 
    \item \textbf{Scene Node Encoding.} We define our holistic scene code based on a unified encoding of walls and object bounding box. Each object $o_j$ is treated as a node with various attributes, i.e., center location $l_i \in \mathbb R^3$, size $s_i \in \mathbb R^3$, orientation $r_i \in \mathbb R$, class label $c_i \in \mathbb R^{C}$. The orientated bounding box is off-the-shelf, we extract the inner walls based on the line junctions and corners of the 3D room. Then we put the orientated object bounding boxes and walls into a compact scene code. Concretely, we define an additional 'empty' object and pad it into scenes to have a fixed number of object across scenes. Each object rotation angle is parametrized by a 2-d vector of cosine and sine values. Finally, each node is characterized by the concatenation of these attributes as $\mathbf o_i = [c_i, l_i, s_i, {\rm cos}r_i, {\rm sin}r_i]$. 
    \item \textbf{data filtering.} We start by filtering out those problematic scenes such as rooms with wall number less than $4$ or larger than $20$. We also remove those scenes with too few or too many objects. The number of walls of valid bedrooms is between $4$ and $10$, and that of objects is between $3$ and $13$. As for living rooms, the minimum and maximum numbers of walls are set to $4$ and $20$, and that of objects are set to $3$ and $24$ respectively. The number of walls for valid kitchens, studies, and bathrooms is the same as for bedrooms, while the objects number is between $3$ and $24$. Thus, the number of scene nodes is $N = 23$ in bedrooms, $N = 44$ in living rooms, and $N = 34$ in kitchens, studies, and bathrooms. After filtering, we get 4,961 bedrooms, 3,039 living rooms, 1,848 kitchens, 638 studies, and 1,500 bathrooms.
\end{itemize}

\noindent \textbf{Text Prompt Generation} We follow the SceneFormer~\cite{wang2021sceneformer} to generate text prompts describing partial scene configurations. Each text prompt contains two to four sentences. The first sentence describes how many walls are in the room, then the second sentence describes two or three existing furniture in the room. The following sentences mainly describe the spatial relations among the furniture, please refer to SceneFormer~\cite{wang2021sceneformer} and DiffuScene~\cite{tang2023diffuscene} for more detailed explanation of relation-describing sentences. In this way, we can get some relation-describing sentences to depict the partial scene. Finally, we randomly sampled zero to two relation-describing sentences to form the text prompt for 3D room generation.

\section{Implementation details}
\label{sec:append:impl}

\noindent \textbf{Training and inference details.} 
\begin{itemize}
    \item In the layout generation stage, We train the scene code diffusion model
on our processed typical indoor rooms data of Structured3D~\cite{zheng2020structured3d} for $200,000$ steps. The frozen text encoder we adopted is the same as Stable Diffusion~\cite{rombach2022high}. The training is performed using the AdamW optimizer with a batch size of $128$ and a learning rate of $1e-4$, utilizing $2$ A6000 GPUs. During the inference process, we utilize the DDIM~\cite{song2020denoising} sampler with a step size of $200$ to perform scene code denoising.
    \item In the appearance generation stage, we fine-tune the segmentation-conditional ControlNet model based on the pairwise semantic and RGB panorama of Structured3D. The fine-tuning process is implemented on two A6000 GPUs for $150$ epochs(about 3 days). In the inference phase, we generate high-fidelity and loop-consistent RGB panorama through DDIM sampler with $100$ steps, rotating both semantic layout panorama and the denoised image for $\mathbf \gamma = 90^\circ$ at each step. 
    \item As for the layout-guided panoramic NeRF module, we set $w_d = 0.6$ and $w_n = 0.4$ for depth alignment loss. During the NeRF fitting process, we randomly select 8 viewpoints for living room scenarios and 4 viewpoints for other room types. The NeRF training settings are the same as PeRF~\cite{wang2023perf}.
    \item As for the mask-guided editing module, we utilize the fine-tuned Control-Seg model to inpaint the background content and optimize the latents of the edited panorama. In inpainting step, the weights used too fuse the unpainted area and unchanged area are set $\lambda_{\rm ori}=0.8,\lambda_{\rm new}=0.2$ . In the optimization step, the maximum iteration is $M=50$, the learning rate $\eta$ for optimization is initialized to $0.1$ and then gradually decreases to $0.01$.
\end{itemize}


\subsection{Baseline Implementations}
\label{sec:append:impl_baselines}

We provide implementation details for baseline methods in the following:
\begin{itemize}
    \item MVDiffusion~\cite{tang2023mvdiffusion}: To get a high-resolution photo realistic panorama, MVDiffusion employs $8$ branches of SD~\cite{rombach2022high} model and correspondence-aware attention mechanism to generate multi-view images simultaneously. We first fine-tune the pre-trained model of MVDiffusion on Structured3D for 10 epochs(about 3 days). Since each generated subview image of MVDiffusion is at $512 \times 512$ resolution, the final panorama is pretty large. We resize the generated panorama of MVDiffusion from $4096 \times 2048$ to $1024 \times 512$. Then the 8 subview perspective images are extracted from the post-processed panorama using the same camera settings (FOV=$90^\circ$,rotation=$45^\circ$). The same operation is adopted on our generated panoramic images. Finally, we combine the panorama from MVDiffusion with the depth estimation~\cite{yun2023egformer} and Poisson reconstruction~\cite{kazhdan2006poisson} module to create a 3D mesh.
    \item Text2Light~\cite{chen2022text2light}: Text2Light creates HDR panoramic images from text using a multi-stage auto-regressive generative model. We choose Text2Light as one of the baseline for our panorama generation and 3D room mesh generation. We first generate RGB panoramas from the input text using Text2Light, then lift it into 3D mesh using the same panoramic reconstruction module as MVDiffusion. When evaluating the panoramic image quality, we adopt the same processing as MVDiffusion to get multi-view perspective images of Text2Light. 
    \item Text2Room~\cite{hollein2023text2room}: Text2Room is the current state-of-the-art and off-the-shelf method for 3D room mesh generation. It utilizes 20 camera spots of a pre-defined trajectory to expand new areas as much as possible by generating 10 images at each spot. Here We use its final fused poison mesh for 3D mesh comparison. For a fair comparison of 2D renderings evaluation, we only use the renderings at the origin of the final mesh.
    
    \item Text2NeRF~\cite{zhang2023text2nerf} generates 3D scenes from a text prompt using NeRF as the 3D representation and leverages a pre-trained text-to-image diffusion model and monocular depth estimation to constrain the 3D reconstruction. However, we found it fails to reconstruct $360^\circ$ scenes. We present some NeRF reconstructions from Text2NeRF stitched into panorama images in Fig.\ref{fig:text2nerf}. Note that only $\sim$154{\textdegree} horizontal field of view (FOV) and $\sim$113{\textdegree} vertical FOV is shown since the rest of the scene is not reconstructed by the method. Thus we skip the comparison with this method.
\end{itemize}

To ensure a fair comparison, we render 60 perspective images at the origin using the final textured meshes of all methods. The camera field of view is set to $140^\circ$ to capture scene layout information for evaluating the CS and IS scores. Additionally, we render corresponding geometric images in \myFigRef{fig:mesh_total} to showcase the geometry quality.

\section{Panorama Generation Comparison}
\label{sec:append:additional_pano_gen_results}

\begin{figure*}[!t]
    \vspace{-0.2in}
    \centering
    \includegraphics[width=1.0\linewidth]{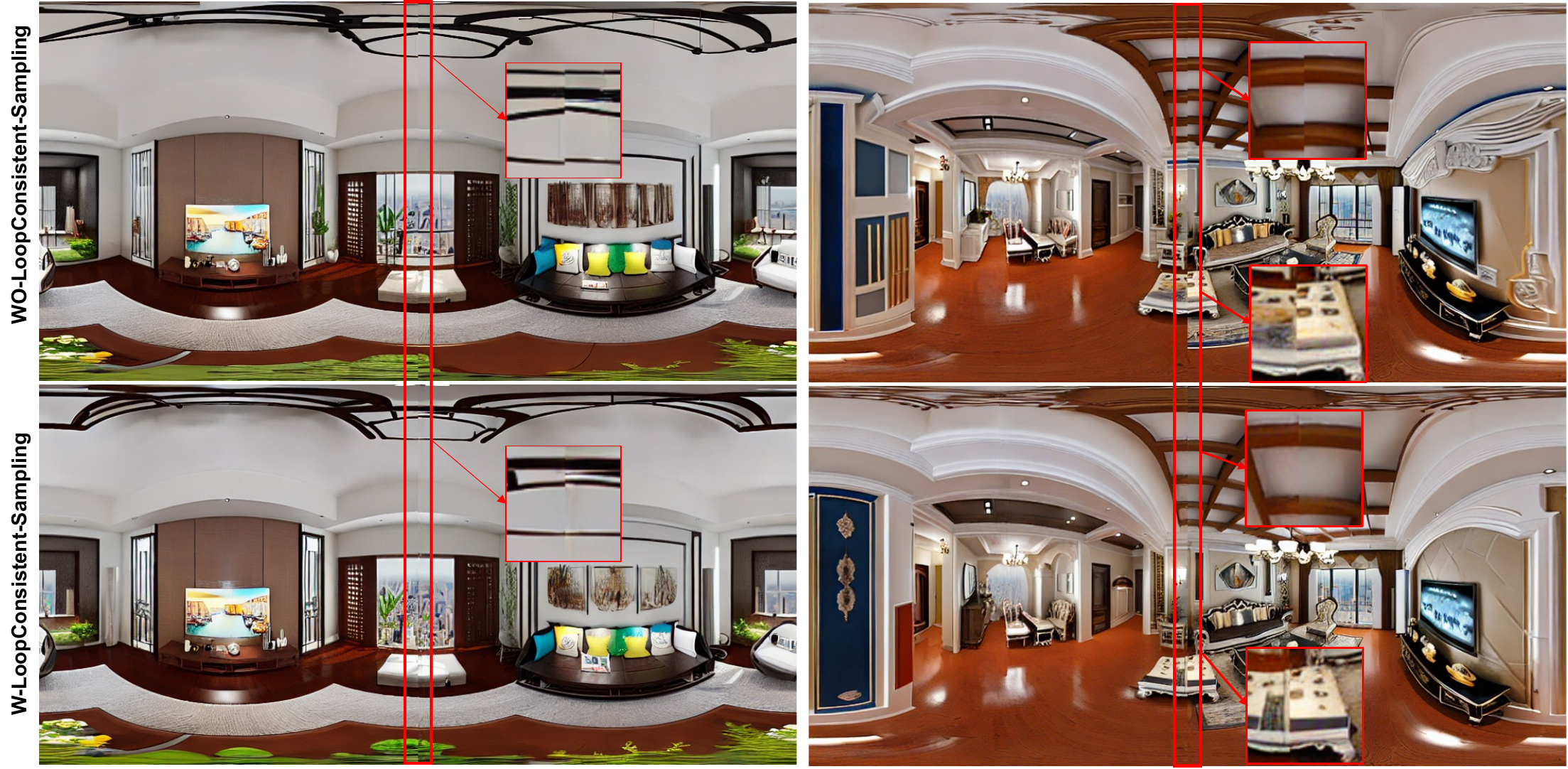}
    \caption{Ablation of loop-consistent sampling examples. We rotate the generated panorama by $180^\circ$ to better visualize the leftmost and rightmost content consistency.  }
    \label{fig:ablation_loop_consist}
    \vspace{-0.2in}
\end{figure*}

\begin{figure*}[t]
    \vspace{-0.2in}
    \centering
    \includegraphics[width=1.0\linewidth]{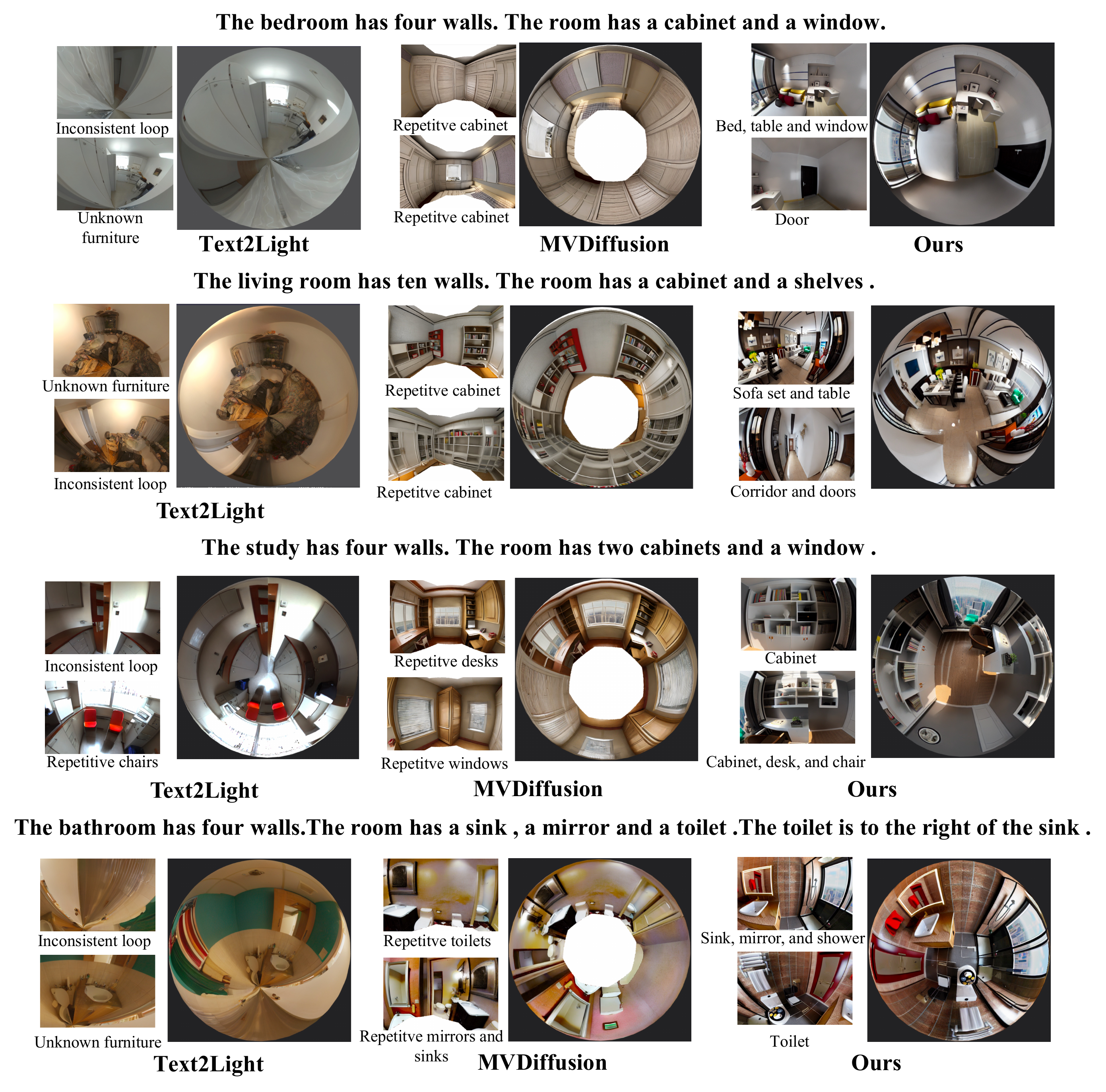}
    \caption{Qualitative comparison for panorama generation. Generated panorama is visualized in a panoramic image viewer to facilitate the user to check the global content of panorama. The left side of each column is two zoom-in views, and the right side is the fisheye view. Text2Light~\cite{chen2022text2light} exists serious inconsistent problem on the border of the generated panorama, it also shows a lot of unexpected stuff in the image. MVDiffusion~\cite{tang2023mvdiffusion} fails to synthesize reasonable content for the target room type. In contrast, our method obtains layout plausible and vivid panorama from the given text prompt of partial scene.}
    \label{fig:append:pano_visualization}
    \vspace{-0.2in}
\end{figure*}

In \myFigRef{fig:ablation_loop_consist}, we study the performance of our panorama generation module with and without loop-consistent sampling mechanism, the ablation indicates the loop-consistent sampling helps the generated panorama obtain better texture consistency.
\myFigRef{fig:append:pano_visualization} presents additional results for panorama generation. Given a simple partial-scene text prompt, our approach obtains better RGB panorama than that of Text2Light~\cite{chen2022text2light} and MVDiffusion~\cite{tang2023mvdiffusion}, which demonstrates the effectiveness of our well-designed framework. While Text2Light suffers from the inconsistent loop and unexpected content of the generated panorama, MVDiffusion fails to recover a reasonable room layout from the text prompt.

\section{Additional Qualitative Results}
\label{sec:append:additional_qual_results}

\begin{figure*}
    \vspace{-0.1in}
    \centering
    \includegraphics[width=1.0\linewidth]{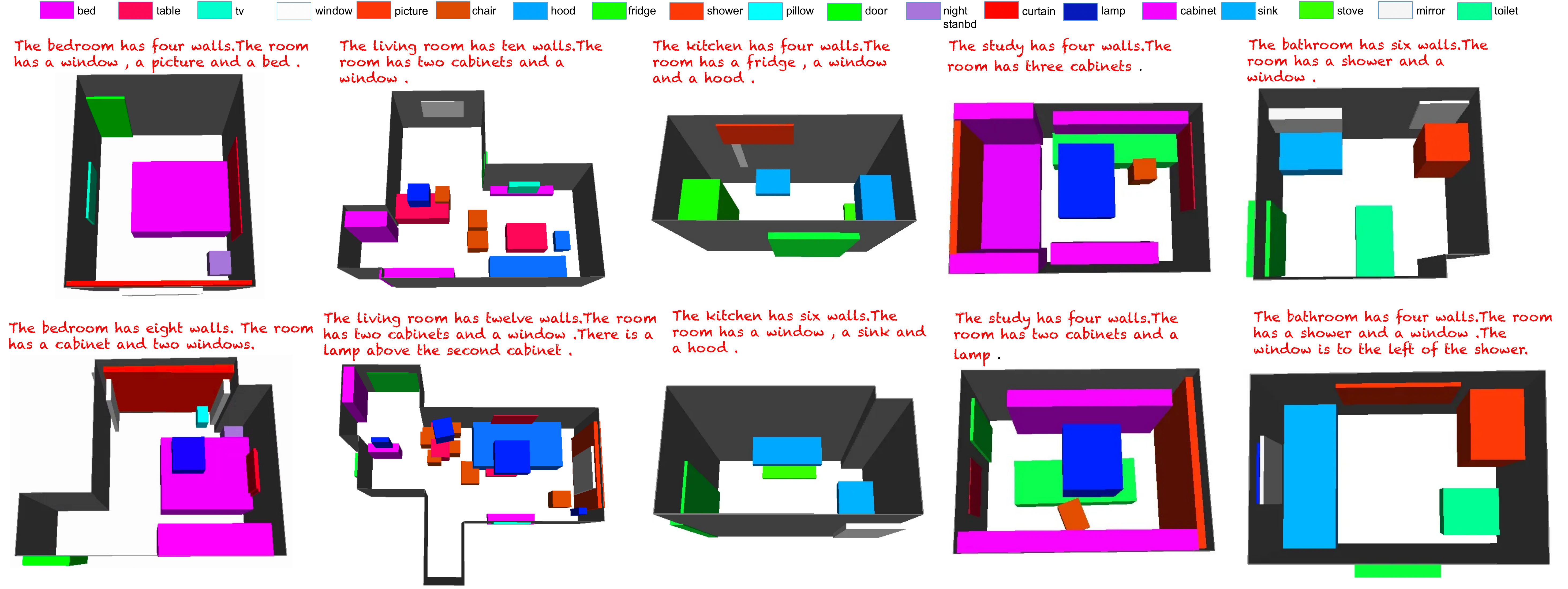}
    \caption{Additional room layout generations. In the bedroom, the bed is often attached to the wall, with a picture above it and a television in front of it. In the living room, there is often a double-seat sofa accompanied by a table and a single-seat sofa. The dining table is usually placed in a separate area of the living room, along with cabinets and chairs. In the kitchen, common furniture includes a stove, sink, fridge, and hood, which are all well-placed in the room. In the study, there is typically a desk accompanied by a chair and one or more bookshelves, and sometimes there is also a bed in the room. In the bathroom, there is usually a sink with a mirror, a toilet, and a shower.}
    \label{fig:append_layout_gen}
    \vspace{-0.1in}
\end{figure*}

\begin{figure*}
    \vspace{-0.1in}
    \centering
    \includegraphics[width=0.85\linewidth]{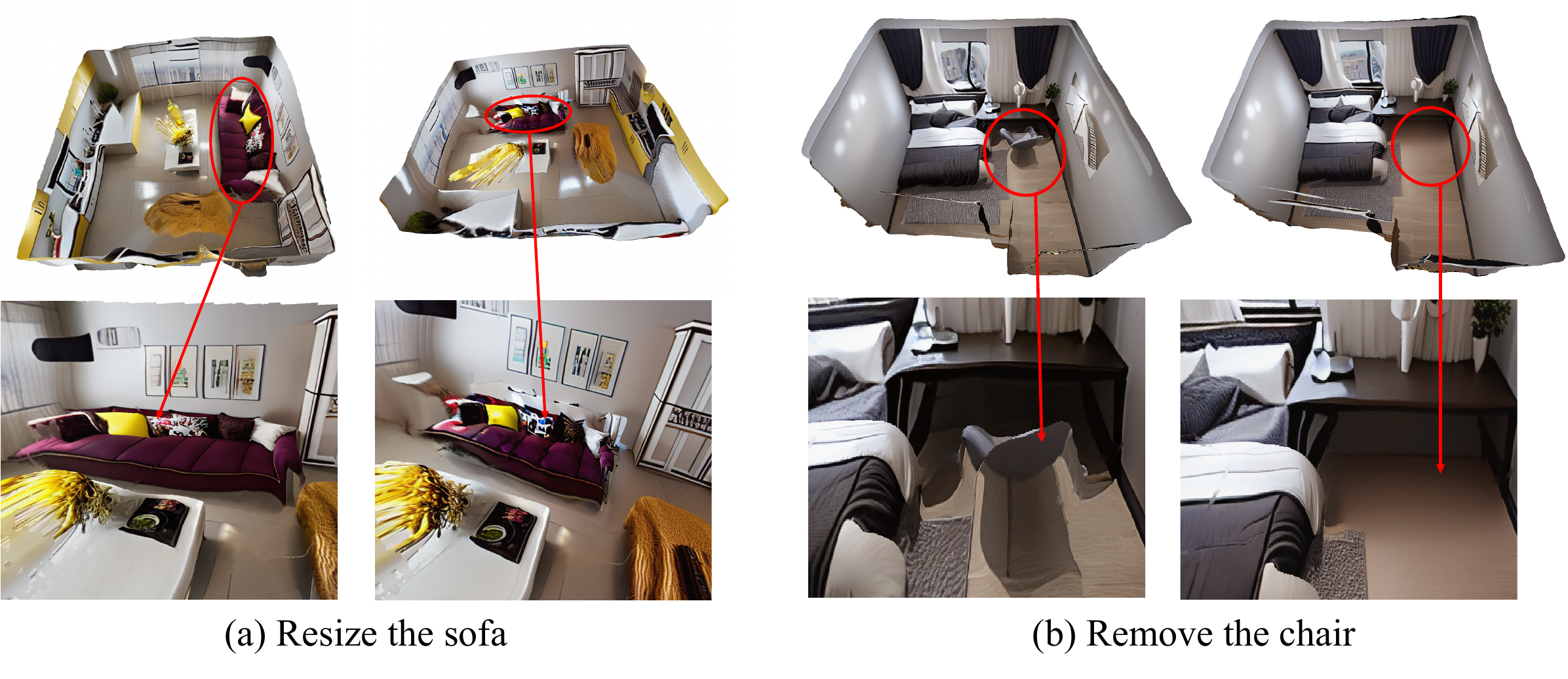}
    \caption{Additional scene editing results. In each sub-figure, the left part is the original 3D room, the right part shows the final mesh after users' interactive editing.}
    \label{fig:append_scene_editing}
    \vspace{-0.1in}
\end{figure*}

\begin{figure*}
    \centering
    \vspace{-0.5in}
    \includegraphics[width=0.95\linewidth]{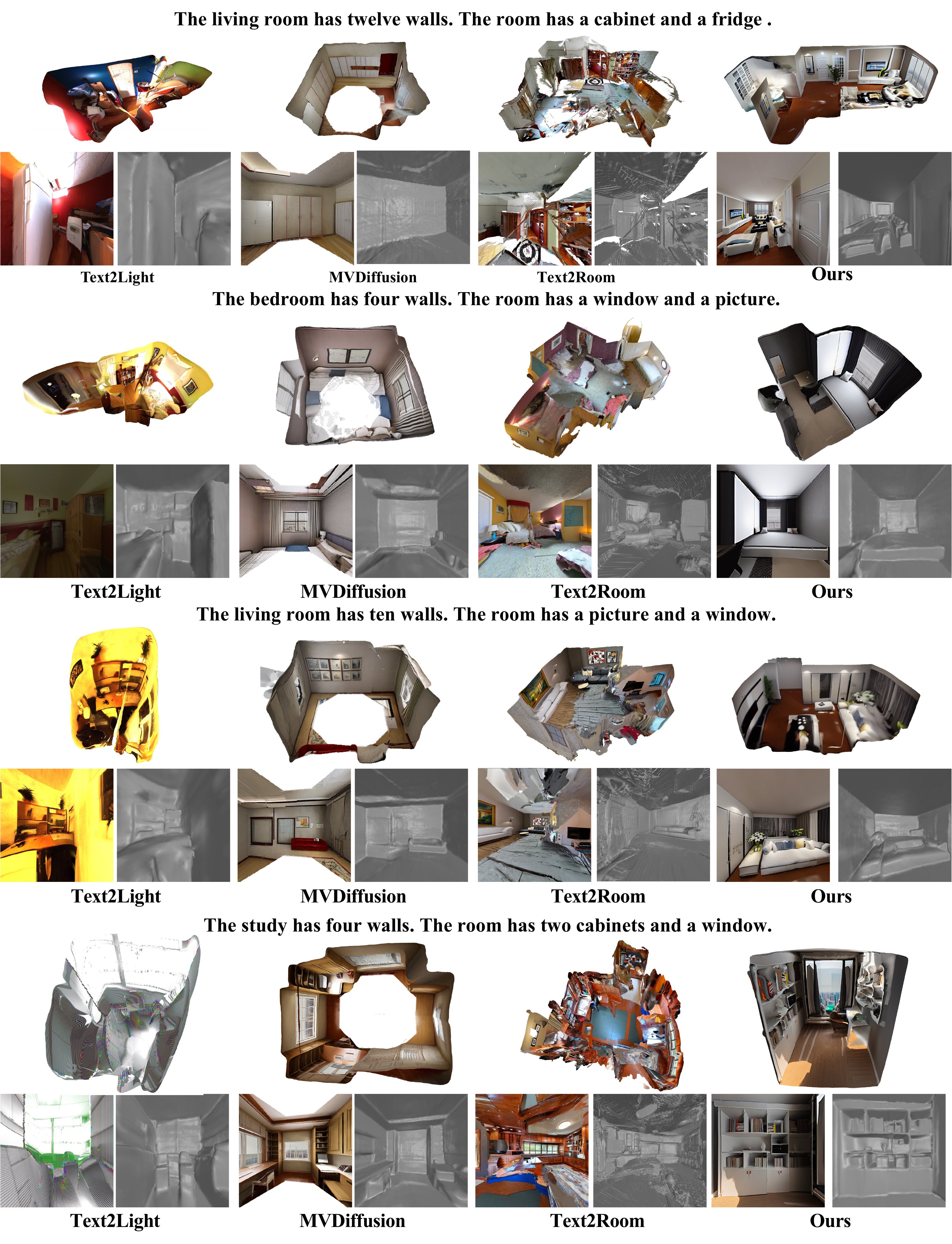}
    \vspace{-0.2in}
    \caption{Additional qualitative comparison with previous works.The first row shows a textured 3D room model, and the second row shows perspective colored renderings and geometric renderings from the room model.    }
    \label{fig:mesh_total}
    \vspace{-0.2in}
\end{figure*}

In \myFigRef{fig:append_layout_gen}, we first visualize more generated room layouts generation of typical rooms in the format of semantic 3D bounding boxes. Then, we show additional qualitative comparison results between our method and baselines in \myFigRef{fig:mesh_total}. We demonstrate more scene editing results of our method in \myFigRef{fig:append_scene_editing}.

\section{User Study}
\label{sec:append:user_study}

\begin{figure*}
    \centering
    \vspace{-0.1in}
    \includegraphics[width=0.85\linewidth]{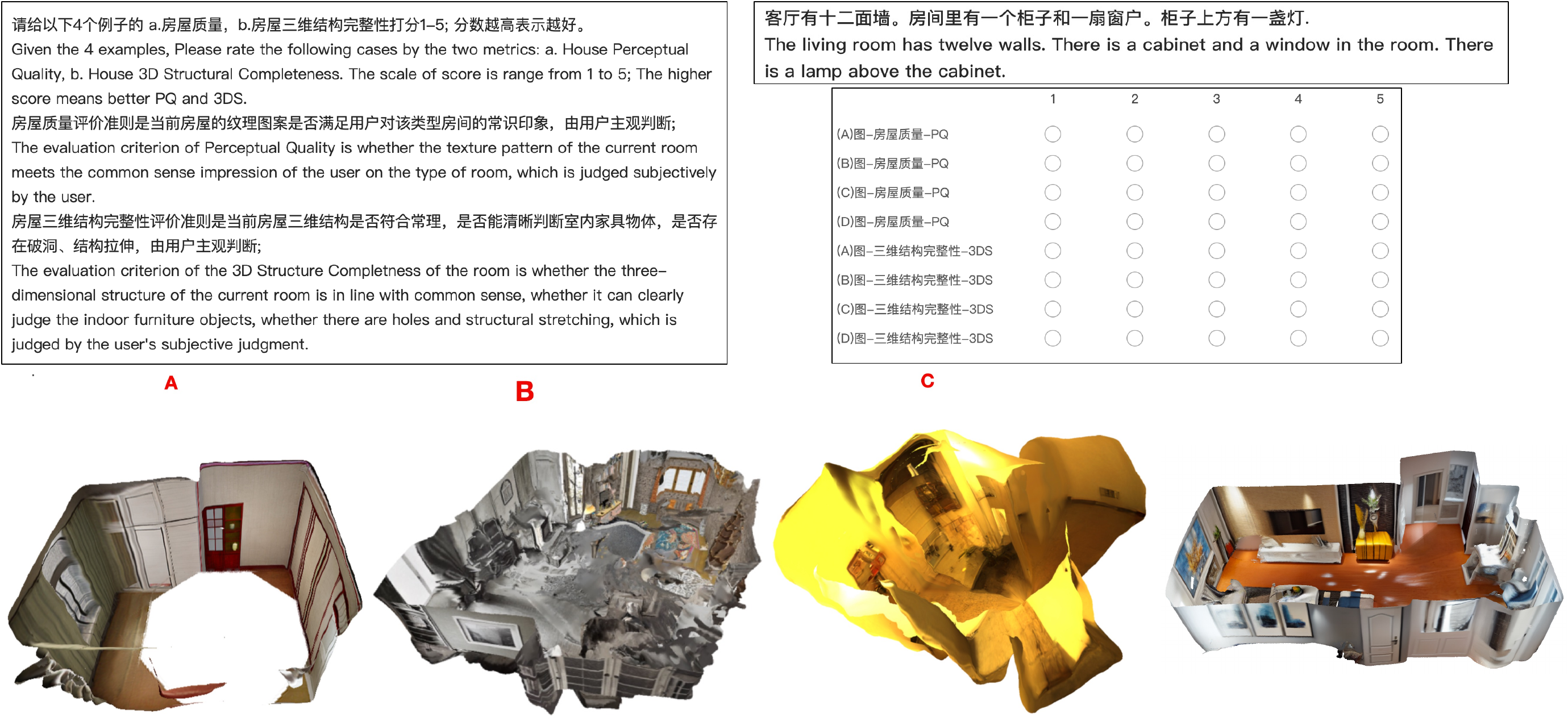}
    \caption{User study interface. We provide users with multiple top-down images from different methods and ask users to rate the given 3D meshes on a scale from 1 to 5, according to the criteria of Perceptual Quality and 3D Structure Completeness.
    }
    \label{fig:append_user_study}
    \vspace{-0.1in}
\end{figure*}

Follow Text2Room~\cite{hollein2023text2room}, we conduct a user study and ask $n=61$ ordinary users to score the Perceptual Quality(PQ) and 3D Structure Completeness(3DS) of the generated room on a scale of $1-5$. Different from Text2room which only demonstrates the perspective renderings of the 3D room, we directly show users the generated mesh to get a global evaluation of the whole generated 3D room. We show an example of the presented interface of the user study in \myFigRef{fig:append_user_study}. In total, we presented $40$ top-down views from $10$ scenes and report averaged results for each method. Users favor our approach, which emphasizes the superiority of our more plausible geometry, along with the vivid texture.

\section{Free style prompts}
\label{sec:append:free_style_prompt}

We show the adaptability of our method by utilizing Large Language Model (LLM) GPT-4 Vision (GPT-4V)~\cite{yang2023dawn} to generate text captions from panorama images of Structured3D~\cite{zheng2020structured3d} bedroom scenes. The prompt used for the LLM is as shown in Table~\ref{table:chatgpt}.

\begin{figure*}[t]
    \vspace{-0.1in}
    \centering
    \includegraphics[width=0.85\linewidth]{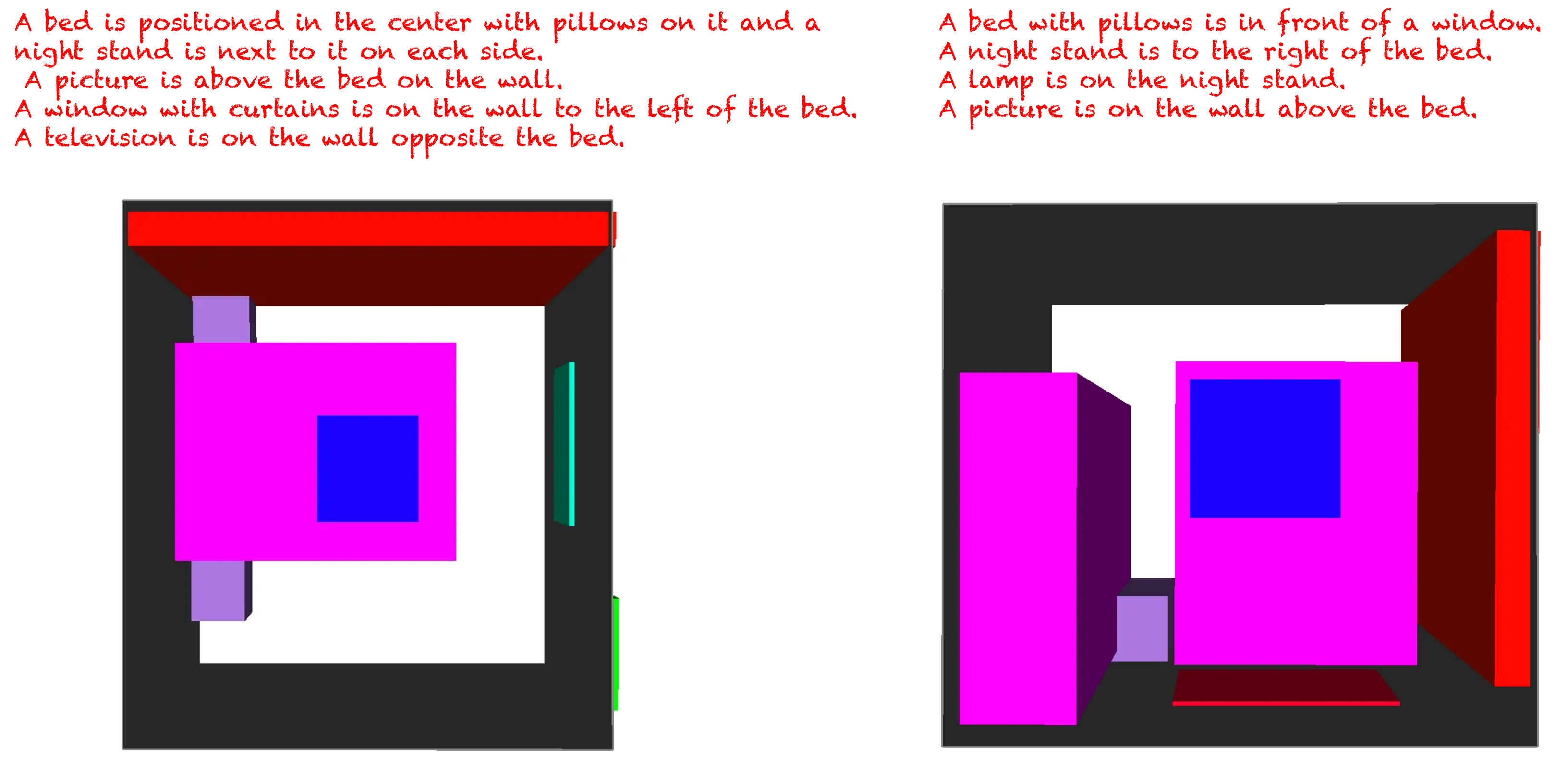}
    \caption{Text-conditioned layout generation on Structured3D using GPT-4V text prompts. Our method synthesizes a plausible scene layout that matches the description.
    }
    \label{fig:exp_gpt4}
    \vspace{-0.1in}
\end{figure*}

\begin{figure*}[t]
    \centering
    \vspace{-0.1in}
    \includegraphics[width=\linewidth]{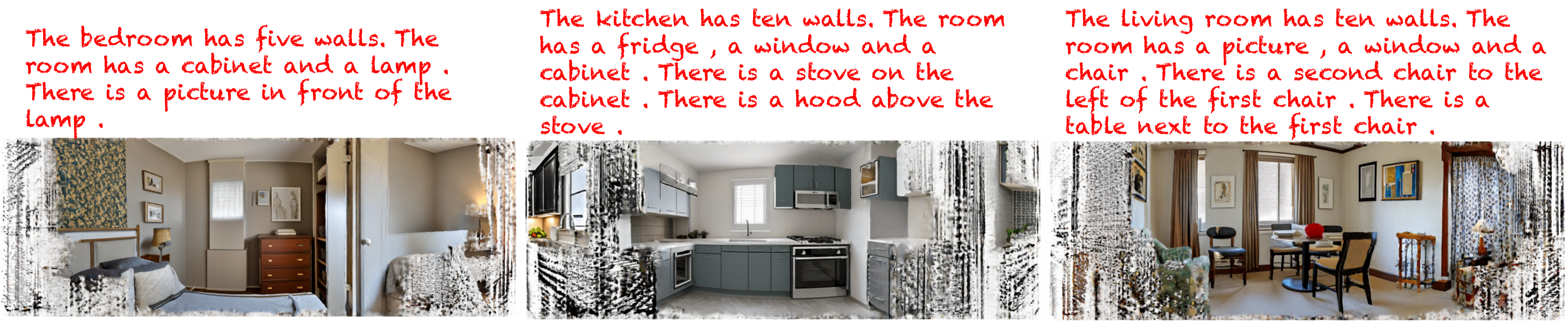}
    \caption{Text2NeRF results. The NeRF reconstructions are stitched into panorama images. Only 154{\textdegree} horizontal FOV and 113{\textdegree} vertical FOV is shown since the method was not able to reconstruct the rest of the scene.}
    \label{fig:text2nerf}
    \vspace{-0.1in}
\end{figure*}

We train and test with the LLM generated captions as conditioning for layout generation. Fig.~\ref{fig:exp_gpt4} shows some results from the test set and corroborates our ability to produce plausible 3D room layout following free-style test prompts.

\begin{table*}[t]
    \centering
    \caption{Prompt for GPT-4V to generate captions from panorama images}
    \label{table:chatgpt}
    \scalebox{0.80}{
    \resizebox{\textwidth}{!}{
        \begin{tabular}{p{15cm}}
            \toprule[1.2pt]
            Describe what is displayed in the panoramic image succinctly in 3 or 4 sentences encoded in ASCII.\\
            Do not use lengthy or compound sentences. Do not mention that it is an image or a panoramic image. \\
            Do not describe the background, lighting, color palette or count the number of objects. \\
            Do not describe size like ``small'', ``large'', etc. \\
            Describe the relative positions of each objects in the scene using only these relationships: ``on'', ``above'', ``surrounding'', ``inside'', ``left touching'', ``right of'', ``front touching'', `in front of'', ``right touching'', ``left of'', ``behind touching'', ``behind'', ``next to'',
            ``left of'', ``right of''. Optionally, describe the object attributes (color, texture etc). \\
            In the description only use these objects: table, night stand, picture, door, cabinet, curtain, bathtub, bed, sink, fridge, shelves, window, lamp, chair, pillow, dresser, bookshelf, sofa, counter, desk, mirror, television, wall
        \end{tabular}
    }}
\end{table*}


{
    \normalem
    \small
    \bibliographystyle{ieeenat_fullname}
    \bibliography{main}
}

\end{document}